# Feasibility of Embodied Dynamics Based Bayesian Learning for Continuous Pursuit Motion Control of Assistive Mobile Robots in the Built Environment

Xiaoshan Zhou[1], Carol C. Menassa[2], and Vineet R. Kamat[3]

[1]Ph.D. Candidate, Dept. of Civil and Environmental Engineering, University of Michigan, Ann Arbor, MI, 48109-2125. Email: xszhou@umich.edu

[2]Professor, Dept. of Civil and Environmental Engineering, University of Michigan, Ann Arbor, MI, 48109-2125. Email: menassa@umich.edu

[3]Professor, Dept. of Civil and Environmental Engineering, University of Michigan, Ann Arbor, MI, 48109-2125. Email: vkamat@umich.edu

**Abstract**

Non-invasive electroencephalography (EEG)-based brain–computer interfaces (BCIs) offer an intuitive means for individuals with severe motor impairments to independently operate assistive robotic wheelchairs and navigate built environments. Despite considerable progress in BCI research, most current motion control systems are limited to discrete commands, rather than supporting continuous pursuit, where users can freely adjust speed and direction in real time. Such natural mobility control is, however, essential for wheelchair users to navigate complex public spaces, such as transit stations, airports, hospitals, and indoor corridors, to interact socially with the dynamic populations with agility, and to move flexibly and comfortably as autonomous driving is refined to allow movement at will. In this study, we address the gap of continuous pursuit motion control in BCIs by proposing and validating a brain-inspired Bayesian inference framework, where embodied dynamics in acceleration-based motor representations are decoded. This approach contrasts with conventional kinematics-level decoding and deep learning-based methods. Using a public dataset with sixteen hours of EEG recordings from four subjects performing a motor imagery-based target-following task, we demonstrate that our method, utilizing Automatic Relevance Determination for feature selection and continual online learning, reduces the normalized mean squared error between predicted and true velocities by 72% compared to autoregressive and EEGNet-based methods in a session-accumulative transfer learning setting. Theoretically, these findings empirically support embodied cognition theory and reveal the brain's intrinsic motor control dynamics in an embodied and predictive nature. Practically, grounding EEG decoding in the same dynamical principles that govern biological motion offers a promising path toward more stable and intuitive BCI control.

## 1 Introduction

In the pursuit of a more inclusive society and the design of accessible public spaces, individuals with limited mobility, including older adults and persons with disabilities, should be afforded comprehensive opportunities to access diverse resources, seek employment, and participate in social gatherings (Belkacem et al., 2020). However, the current reality is that restricted mobility often limits these individuals' ability to benefit fully from various social welfare provisions (Bezyak et al., 2019; Jansuwan et al., 2013; NEVEN, 2015). This constraint extends even to essential daily activities such as shopping for necessities and obtaining medical care.

This issue is both widespread and urgent, affecting a substantial portion of the population worldwide. Rapid population aging and the rising prevalence of disabilities have created an urgent need for assistive technologies that support independence and dignity in daily life (Remillard et al., 2022). For instance, in the United States alone, demographic projections indicate significant shifts: by 2030, one in five people will be aged 65 or older, and by 2034, older adults are expected to outnumber children under 18 (AARP, 2024). Concurrently, nearly one in four adults—over 70 million people—live with some form of disability (CDC, 2024). These statistics highlight the pressing necessity for innovative technologies that help enhance independent mobility, reduce reliance on caregivers (Al-Qaysi et al., 2018; Davies et al., 2003), and enable more equitable participation in social and economic activities.

At present, people with mobility impairments predominantly rely on wheelchairs for daily movement, which may be manually operated or powered by electric motors. The advent of motorized wheelchairs has significantly alleviated the physical effort required for manual propulsion (Davies et al., 2003), while also expanding the range and diversity of activities accessible to users. For example, motorized wheelchair users can now navigate outdoors under adverse weather conditions: traversing snowy surfaces, enduring strong winds, or navigating rain while holding an umbrella. The enhanced mobility has further expanded the functional and socioeconomic opportunities available to wheelchair users. For instance, the ability to transport goods or carry groceries powered by a motorized wheelchair enables individuals who were once isolated at home to actively participate in community engagements and even take on gig-economy roles such as local deliveries. This promotion of meaningful inclusion in society has the potential to change the current reality social isolation and workforce exclusion often faced by wheelchair users (NEVEN, 2015), especially in less developed countries.

Furthermore, robotic wheelchairs equipped with intelligent path planning algorithms extend users' ability to navigate both indoor buildings and outdoor sidewalks efficiently. It also makes it feasible for users to navigate in unfamiliar environments once impossible, for instance, finding their way to a specific examination room during a first visit to a hospital. However, although fully autonomous wheelchairs are technologically advanced, they also raise significant usability concerns. When navigation is governed entirely by autonomous systems, users may feel disempowered or anxious, as their sense of control is diminished and they may feel as though they become "marionettes". This highlights the need for an intermediate approach, i.e., shared control, which maintains the benefits of motorized power and intelligent path planning, yet simultaneously preserve human agency and the ability to adjust or override the wheelchair's operation as needed.

Shared control systems could offer substantial practical advantages for robotic wheelchairs. While autonomous wheelchairs are equipped with obstacle avoidance capabilities, their navigation in real indoor environments remains limited. For example, one of the key challenges is socially aware navigation, including determining the appropriate social distance to maintain from pedestrians or follow others. Although robotic wheelchairs can be programmed to follow general design guidelines, each user's psychological comfort level varies, and individuals may have their own preferences to make adjustments. Additionally, autonomous navigation typically operates by setting a start and end point and executing a route once selected at the beginning. However, in real-world driving scenarios, users may often need to change their navigation intentions. For example, in an airport, a wheelchair user may encounter a crowded situation in front and prefer to stop and wait until congestion clears, rather than proceeding to navigate around the crowd. In a hospital, users might meet acquaintances or medical staff along the way and wish to pause for conversation, or may need to make unscheduled stops to use restroom facilities.

Because of these dynamic interactions and decisions that arise during navigation, we need to understand users' motion intention (desired changes in speed and direction) in real time in order to flexibly accommodate such interventions during autonomous operation. One intuitive approach for enabling this user input is through brain-computer interfaces (BCIs), which have been developed to restore communication and control capabilities for individuals with severe motor impairments (Frank et al., 2000; Luis & Gomez-Gil, 2012). BCIs bypass muscle pathways entirely, directly interpreting brain signals to discern users' intentions for wheelchair control (Bi et al., 2013). Despite extensive research into EEG-based BCI-driven wheelchairs, most control schemes remain limited to discrete commands, such as move forward, stop, turn left or right, and select among predefined destinations or trajectories (Chen et al., 2022; Hamad et al., 2017; He et al., 2017; Iturrate et al., 2009a; Iturrate et al., 2009b; Kim & Lee, 2016; Lakas et al., 2021; Ng et al., 2014; Ngo & Nguyen, 2022; Widyotriatmo et al., 2015; Yu et al., 2017)

The key underlying rationale for the methodological framework developed in this study is to decode brain activity in a manner consistent with how humans generate intention-encoded signals, specifically through a Bayesian inference approach (Friston, 2010; Knill & Pouget, 2004; Thomas Parr, 2022). Moreover, prior research typically focuses on decoding Electroencephalography (EEG) signals and mapping them to velocity or position to control wheelchairs (Forenzo et al., 2024b; Inoue et al., 2018; Lee et al., 2025; Volkova et al., 2019; Willsey et al., 2025). However, the human brain fundamentally organizes force to drive motion; therefore, we hypothesize that motion control is more appropriately represented at the dynamics level, rather than the kinematics level. This paper thus aims to evaluate a Bayesian learning approach from an embodied dynamics perspective.

Section 2 provides a detailed review of state-of-the-art EEG-based BCI-driven wheelchair systems and outlines the neuroscience inspirations underpinning our Bayesian methodology. To assess our method's performance, particularly its capability for continual online learning during use for reliable BCI control, we tested it under two control paradigms (kinematics versus dynamics decoding) using a public continuous pursuit BCI dataset. Sixteen hours of continuously recorded EEG data (acquired with a 64-channel Neuroscan EEG cap) and corresponding kinematics performance from four users were analyzed in a real-time, motor imagery (MI)-based target-following task. Automatic Relevance Determination (ARD) was used to select meaningful channels and synchronized features across four frequency subbands of the EEG signals. We compared the error between predicted and true velocities achieved by our method with four widely used training strategies in recent literature: an autoregressive model, EEGNet-based deep learning, transfer learning, and calibrated learning. Finally, we propose a feasible neurophysiological mechanism for BCI-enabled motion control, and discuss its broader implications for embodied artificial intelligence (AI) systems.

## 2 Related Works

### 2.1 Interaction Modalities for Shared Control in Robotic Wheelchair

Contemporary research on robotic wheelchairs has explored a spectrum of interaction modalities to facilitate shared control, aiming to understand human intention to enable human input, particularly for individuals with severe motor impairments. Commonly implemented modalities include voice commands (Williams & Scheutz, 2017) and vision-based gesture recognition (Prathibanandhi et al., 2022). These approaches can be effective in controlled, quiet environments, such as private residences, where ambient noise is minimal and social pressures are limited (Al-Qaysi et al., 2018). In such settings, speech and gesture recognition systems typically perform well, enabling hands-free or physically unobtrusive control.

However, these modalities reveal critical shortcomings in real-world, public environments. The performance of voice-based interfaces deteriorates markedly in noisy settings, such as transit stations or busy streets, where high background noise introduces recognition errors and reduces reliability. Likewise, vision-based gesture systems face substantial challenges due to variable lighting conditions, complex backgrounds, and line-of-sight occlusions. Thus, their effectiveness is confined to relatively static and predictable environments, limiting their practical utility for users who need robust, context-independent interface solutions.

EEG-based BCI control has emerged as a promising alternative, offering distinct advantages that address many of these limitations. Most notably, EEG-based systems bypass peripheral muscular pathways and directly capture the neural signatures of movement intention (Maiseli et al., 2023). By reading brain activity rather than observable actions or auditory cues, EEG-driven interfaces remain largely immune to environmental disturbances such as noise, visual clutter, or physical barriers, enabling reliable performance in diverse and dynamic settings where conventional methods often fail.

Beyond technical robustness, EEG-driven control also delivers critical psychological and social benefits. Because neural activity can be interpreted silently and invisibly, EEG-based interaction provides a private and discreet channel of communication. Users need not vocalize instructions or perform visible gestures, preserving confidentiality and reducing the risk of social discomfort or unwanted attention in public (Freitas et al., 2017). In contrast, speech-based systems can inadvertently broadcast user intentions to bystanders, while gesture-based controls may be conspicuous or misunderstood. For disabled individuals whose sense of autonomy and dignity may be especially vulnerable in social contexts (Bezyak et al., 2019), such privacy considerations are not trivial.

## 2.2 EEG-Based BCI-Driven Wheelchair Motion Control

Two dominant paradigms have shaped the landscape of EEG-based BCI research: spontaneous and evoked signal paradigms. Spontaneous signals stem from internal mental activities, most notably MI, in which users imagine specific movements such as left or right hand motion (Cao et al., 2014; Huang et al., 2019) or foot movement (Li et al., 2013; Long et al., 2012). This paradigm forms the basis for many early BCI competitions and public benchmark datasets (Blankertz et al., 2006; Gwon et al., 2023). In addition to MI, other cognitive imagery tasks, such as mental singing, arithmetic, and spatial rotation, have also demonstrated distinct and classifiable neural patterns (Hwang et al., 2014). Evoked signals, on the other hand, are generated in direct response to external stimuli, with the oddball P300 (He et al., 2017; Iturrate et al., 2009a; Yu et al., 2017) and steady-state visual evoked potentials (SSVEP) (Chen et al., 2022; Kim & Lee, 2016; Long et al., 2012; Widyotriatmo et al., 2015) being the most well-studied examples.

Despite continuous advancement within these paradigms, a critical gap persists in their practical application to mobile robot control, such as wheelchair navigation. Both spontaneous and evoked paradigms were originally designed for discrete, low-dimensional tasks, such as selecting characters or initiating simple predefined actions. They do not naturally translate to the seamless, continuous, and intuitive control needed for real-world movement. Consequently, bridging the gap between laboratory-based BCI paradigms and embodied, fluid robotic interaction remains an open challenge.

A central technical obstacle is the non-stationarity of EEG signals. Minor electrode shifts, changes in skin impedance, and fluctuations in user attention, fatigue, or emotional state can produce substantial signal drift, undermining decoding accuracy over time (Silversmith et al., 2021; Zhang et al., 2020). Existing solutions have predominantly relied on transfer learning and continual learning strategies (Fahimi et al., 2019; Forenzo et al., 2024b; Zhang et al., 2020; Zhou & Liao, 2023), primarily leveraging deep neural network frameworks that attempt to recalibrate model weights as the underlying data evolves. While deep learning methods have demonstrated notable adaptive capabilities, a fundamental question persists: to what extent do artificial neural networks actually mirror the brain's computational principles, and are they truly optimal for decoding neural signals in MI and other tasks? The conceptual analogy between artificial neural networks and biological brains remains largely superficial, given the profound differences in their learning mechanisms (Plebe & Grasso, 2019).

This discrepancy highlights the importance of exploring alternative inference architectures. In particular, active inference, grounded in the Bayesian brain hypothesis, offers a theoretically robust and biologically plausible framework for understanding how humans generate and adapt motor commands in uncertain environments (Friston, 2010; Knill & Pouget, 2004; Thomas Parr, 2022). Although its application to neural decoding remains largely underexplored, active inference offers interpretability and holds promise for advancing BCI-based motor control. Therefore, empirically validating such Bayesian approaches may bridge the gap between biological computation and artificial systems, illuminating new paths for BCI research.

A second major challenge in the field is that current BCI-controlled mobile robots continue to depend heavily on discrete control paradigms. In most existing systems, once MI patterns are decoded from EEG signals (Abiyev et al., 2016; Fernández-Rodríguez et al., 2016; Li et al., 2013; Li et al., 2025; Tabar & Halici, 2017; Wang et al., 2014), they are mapped to a limited set of predefined commands or selections. These include character selection in BCI spellers (Blankertz et al., 2006; Candrea et al., 2024; Gwon et al., 2023), destination selection for navigation (Iturrate et al., 2009a; Iturrate et al., 2009b; Lakas et al., 2021; Ng et al., 2014; Ngo & Nguyen, 2022), trajectory confirmation (Hamad et al., 2017), or simple control commands such as turning left or right, moving forward, and stopping (Chen et al., 2022; He et al., 2017; Kim & Lee, 2016; Widyotriatmo et al., 2015; Yu et al., 2017).

However, this framework falls short of the ideal paradigm for BCI-based assistive robotic wheelchair, which should enable smooth, flexible, and continuous interaction. In an optimal system, users would be able to steer the robot freely in any direction, refining trajectories and movement in real-time, rather than being restricted to a finite set of discrete options. This research challenge is conceptualized as the "continuous pursuit task" in BCI studies (Forenzo et al., 2024a). Yet, despite extensive advancements using left- and right- hand MI for directional control, the continuous pursuit task where users exert ongoing, fluid control over speed and direction remains largely unaddressed, but it is a critical frontier for advancing truly seamless and natural BCI-driven control of assistive mobile robots in shared control frameworks.

## 2.1 Deep Learning for EEG Motor Imagery Decoding

MI refers to the mental simulation of physical movement without actual muscle execution and represents one of the most widely studied paradigms in BCI research (Altaheri et al., 2023). Over the years, both traditional machine learning (ML) and deep learning methods have been developed to decode MI-related EEG signals. Among traditional ML approaches that rely on handcrafted features, the Common Spatial Pattern (CSP) feature and its variants have consistently demonstrated strong performance in MI-EEG classification (Ang et al., 2012; Xie & Li, 2015). These methods extract discriminative spatial–spectral features by decomposing EEG signals into multiple frequency bands (Millan et al., 2009), which are subsequently fed into classic classifiers such as Linear Discriminant Analysis (LDA) and Support Vector Machines (SVM) (Fernández-Rodríguez et al., 2016).

The emergence of deep learning has transformed EEG decoding by enabling models to automatically learn rich, discriminative representations directly from raw signals, thus eliminating the need for manual feature engineering. This capability is particularly valuable given the inherent complexity of EEG data, which are high-dimensional, non-stationary, and characterized by a low signal-to-noise ratio in the temporal domain (Hasenstab et al., 2017). Researchers have explored a wide range of neural architectures for MI-EEG decoding, including autoencoders (AEs) (Hassanpour et al., 2019), convolutional neural networks (CNNs) (Lawhern et al., 2016), recurrent neural networks (RNNs) (Luo et al., 2018), temporal convolutional networks (TCNs) (Ingolfsson et al., 2020), deep belief networks (Xu et al., 2020), and more recently, Transformer-based models (Wan et al., 2023). Among these architectures, CNNs have become the most widely adopted backbone due to their ability to capture localized spatial–temporal patterns in EEG data efficiently (Ang et al., 2012; Saibene et al., 2024).

Both lightweight and deep convolutional CNN architectures have been explored for MI-EEG classification, including a wide range of variants such as inception-based CNNs (Amin et al., 2021), multi-scale CNNs (Li et al., 2020), multi-branch CNNs (Liu & Yang, 2021), and attention-enhanced CNNs (Altuwaijri et al., 2022). Notable baseline models include DeepConvNet and ShallowConvNet proposed by (Schirrmeister et al., 2017), which employ deep and compact convolutional layers to extract hierarchical representations from EEG data. Another widely adopted model is EEGNet by (Lawhern et al., 2016), which leverages depthwise and separable convolutions to efficiently learn temporal and spatial features from multi-channel EEG signals.

Subsequent studies have incorporated temporal modeling components alongside CNNs to better capture sequential dependencies and temporal dynamics, most commonly through RNNs and TCNs. Additionally, attention mechanisms have been integrated to enable models to selectively emphasize task-relevant neural patterns while suppressing noise. Although these hybrid architectures have achieved improved performance on benchmark datasets such as BCI Competition IV-2a and IV-2b (Zhao et al., 2025), EEGNet continues to serve as one of the most robust and representative deep learning baselines in recent real-time, BCI-enabled control studies (Forenzo et al., 2024b; Lee et al., 2025).

## 2.4 Theoretical Foundation

### 2.4.1 Neuroscience-inspired Artificial Intelligence

In the early development of AI, research was deeply intertwined with neuroscience and psychology, as scientists sought to model aspects of human cognition and neural computation (Hinton, 1984; Hopfield, 1982; Turing, 2007). Over time, debates emerged regarding whether the brain should serve as the ideal blueprint for machine intelligence or whether artificial systems should diverge to pursue alternative computational paradigms (Brooks et al., 2012).

In the past decade, AI has undergone a profound transformation driven by the rise of neural network–based, or "deep learning" methods (LeCun et al., 2015). The computational mechanisms underlying CNNs, including nonlinear transduction, divisive normalization, and max-pooling (Yamins & DiCarlo, 2016), were directly inspired by single-cell recordings in the mammalian visual cortex, which revealed how neurons in primary visual area V1 filter and integrate sensory inputs (Hubel & Wiesel, 1959). Moreover, modern network architectures mirror the hierarchical organization of the mammalian cortical system. They implement both convergent and divergent information flow across successive layers, analogous to the hierarchical processing observed for object recognition in biological visual pathways (Krizhevsky et al., 2012; Riesenhuber & Poggio, 1999; Serre et al., 2007).

However, despite the remarkable success of deep learning, its role in advancing toward artificial general intelligence (AGI) remains highly debated. After reviewing 84 cognitive architectures proposed over the past four decades, (Kotseruba & Tsotsos, 2020) concluded that deep learning models, while powerful for pattern recognition, do not constitute unified cognitive architectures and fall short of modeling the full scope of human intelligence. Although artificial neural networks are often described as being inspired by the human brain, several fundamental neural mechanisms, such as spiking dynamics, lateral connectivity, and Hebbian learning, are absent from modern deep learning frameworks. Consequently, deep learning differs substantially from biological neural networks in both design principles and functional scope (Maex et al., 2009). Historically, neural network research has experienced periods of decline and resurgence, with a major turning point marked by the introduction of the backpropagation algorithm, which enabled efficient learning across multiple network layers (Rumelhart et al., 1985). Nevertheless, backpropagation lacks biological plausibility. Unlike artificial networks that rely on a global error signal propagated across all layers, biological neurons update synaptic weights based on local learning rules, where adjustments depend on local activity patterns and synaptic correlations (Plebe & Grasso, 2019; Rolls, 2023). Human neurons do not communicate over long distances to share centralized error information; thus, the mechanism underpinning modern deep learning are fundamentally different the processes governing learning in the brain.

For decoding neural signals, the human brain contains approximately 100 billion neurons interconnected by trillions of synapses (Herculano-Houzel, 2009; Pakkenberg et al., 2003). EEG recordings measure voltage fluctuations at the scalp that represent the summed activity of millions of neurons, primarily the synchronized dendritic currents of pyramidal cells within the cortical neuropil (Buzsáki, 2006). These signals provide a coarse and highly integrated view of neural dynamics, where fine-grained spatiotemporal information can be lost due to volume conduction and spatial averaging. This inherent complexity and indirectness mean that EEG decoding does not map onto the layer-by-layer computations of artificial neural networks. In addition, deep learning architectures assume structured, stationary relationships between inputs and outputs and depend on large, high-quality datasets for stable training. This condition is however rarely satisfied by real-time EEG data, which are noisy, non-stationary, and influenced by transient cognitive and physiological states. Moreover, unlike the biological brain, which performs probabilistic inference over uncertain sensory evidence, neural networks are deterministic function approximators that lack explicit mechanisms for representing uncertainty.

Therefore, although deep learning has achieved impressive performance in EEG signal classification, it may not represent the most suitable approach for decoding EEG signals related to flexible motor control (the underlying neural computations are discussed in Section 2.4.2). This limitation instead motivates the exploration of alternative frameworks, particularly the Bayesian and active inference approaches that more closely reflect how the brain integrates noisy sensory inputs, updates beliefs, and regulates action under uncertainty.

### 2.4.2 Embodied Cognition and Internal Models of Motion

Embodied cognition posits that perception and action are not separable processes but deeply intertwined, with cognition grounded in the bodily structures and sensorimotor capacities of the agent. Unlike traditional representational theories that treat cognition as an abstract manipulation of symbols (Dawson, 2019), embodied accounts argue that the mind is shaped and constrained by the morphology of the body and the dynamics of its interaction with the world. Philosophical perspectives (Shapiro & Spaulding, 2024) frame embodiment as a rejection of Cartesian dualism, reasserting the role of the body, beyond the brain itself, in structuring cognition. The enactive approach further emphasizes that cognition emerges from embodied sensorimotor experience, situated within biological, psychological, and cultural contexts (Varela et al., 1991). Collectively, these accounts that the nervous system's representations are not arbitrary but are tuned to the dynamic variables through which the body acts upon the world. This view naturally supports a force-based interpretation of motor control: since movement is governed by physical mass, motor execution is most plausibly organized at the level of acceleration, consistent with Newton's second law.

In human motion perception, embodiment manifests in anticipatory processes such as representational momentum, which is the systematic displacement of a perceived object position forward along its trajectory (Brouwer et al., 2004; Freyd & Finke, 1984). Such predictive biases indicate that the brain does not passively register positions but instead encodes dynamic internal models that anticipate future motion states. This interpretation aligns with the predictive coding framework (Hohwy, 2013), which proposes that higher-level generative models continuously predict upcoming sensory input. From an embodied perspective, representational momentum can be viewed as a cognitive analogue of physical inertia, implying that the brain encodes not only position but also concepts of mass and force. Therefore, motor control can be understood as organized primarily at the acceleration level as a direct expression of embodied dynamics.

Findings from motor control research further support this interpretation. Human movements exhibit a number of invariant properties, such as Fitts' law (Fitts & Peterson, 1964), the bell-shaped velocity profile (Morasso, 1981), the two-thirds power law (Viviani & Schneider, 1991), and the minimum-jerk model (Hogan, 1982), which are believed to reflect fundamental principles of voluntary motion organization (Harris & Wolpert, 1998; Jones et al., 2002). Among these, the minimum-jerk model has been particularly influential: jerk, the derivative of acceleration, is both directly measurable and strongly predictive of natural human trajectories (Flash & Hogan, 1985; Shadmehr & Wise, 2004; Todorov, 2004). If the nervous system indeed minimizes jerk, producing the smooth, bell-shaped velocity profiles and straight-line trajectories characteristic of natural movement, then the primary control variable is best interpreted as acceleration (the integral of jerk). Velocity and position, in turn, represent secondary, noisier integrations of these underlying dynamics. This perspective motivates our central hypothesis: BCI-based continuous motion control should prioritize decoding at the acceleration level, aligning more closely with the dynamic principles that govern biological motor behavior.

### 2.4.3 Bayesian Inference

The Bayesian brain hypothesis conceptualizes the nervous system as an inferential machine that encodes and computes probability distributions to interpret and predict sensory events (Friston, 2012; Knill & Pouget, 2004; Pouget et al., 2013). Under this view, perception and action are framed as probabilistic inferences performed under uncertainty: because sensory observations are inherently noisy and ambiguous (Beck et al., 2012; Fiser et al., 2010), they must be integrated with prior expectations about environmental regularities (Girshick et al., 2011; Hohwy, 2017; Ma et al., 2006; Quax et al., 2021). Neural circuits are thought to implement generative models that continuously compare predicted states with incoming sensory input, updating beliefs via prediction error minimization (Shadmehr et al., 2010; Wolpert et al., 2011). Through this recursive updating, organisms can approximate Bayes-optimal decision making and action selection (Körding & Wolpert, 2004; Wolpert & Landy, 2012).

Empirical evidence for Bayesian inference in the brain has been observed across multiple domains. Event-related potentials such as mismatch negativity and the P300 have been interpreted as neural signatures of prediction error and Bayesian surprise (Baldi & Itti, 2010; Näätänen et al., 2011; Polich, 2007). EEG studies demonstrate that trial-by-trial activity correlates with Bayesian parameters such as prior expectation and surprise during attention and decision-making tasks (Gómez et al., 2019). Similarly, multisensory integration exemplifies Bayesian computation: rather than simply fusing signals, the nervous system infers the most likely causal structure underlying them (Aller & Noppeney, 2019; Rohe et al., 2019), a process that unfolds hierarchically over time.

Beyond biological cognition, Bayesian inference provides a principled foundation for robotics and AI. It underlies key processes such as state estimation (e.g., particle filtering and Monte Carlo localization; (Li et al., 2024)), affordance modeling (Montesano et al., 2008), gesture recognition (Lopes & Santos-Victor, 2005), and adaptive trajectory planning (Shen et al., 2025). In neuroscience, Bayesian methods are widely applied to EEG for source localization (Jun et al., 2005; Yildiz et al., 2007) and ERP detection (Wu et al., 2014), though typically in offline settings. Only a few studies have attempted to embed Bayesian inference into real-time BCI systems (Chiappa, 2006), and these have not explicitly addressed temporal uncertainty in user intention or integration within closed-loop sensorimotor control.

Taken together, Bayesian inference offers a powerful computational framework for modeling uncertainty, prediction error, and internal generative models—principles fundamental to human motor control yet underexplored in continuous BCI applications. This motivates our investigation into whether applying Bayesian inference to acceleration-level decoding can yield biologically aligned improvements in continuous BCI-based motion control.

## 3 Methodology

### 3.1 Dataset

We evaluate the proposed framework using the Continuous Pursuit (CP) BCI dataset (Forenzo et al., 2024a; Forenzo et al., 2024b), a uniquely publicly available benchmark specifically designed for real-time continuous BCI control.

In this dataset, participants used non-invasive EEG recordings to continuously control a cursor that tracked a moving visual target on a two-dimensional screen. The task is grounded in the left- and right-hand MI paradigm, which is dominant in MI-based BCIs (Saibene et al., 2024). Participants were instructed to employ a kinesthetic MI strategy, i.e., imagining the feeling of hand movement. This is different from the visual imagery paradigm, where participants imagine observing motion without the accompanying sensation. In previous BCI studies, it is reported that the kinesthetic strategy is more reliable than the visual strategy (Xiong et al., 2019).

Control was intuitive: imagining the left hand moved the cursor leftward, the right hand moved it rightward, both hands simultaneously moved it upward, and no hand movement (idle state) moved it downward. Moreover, participants could combine these strategies to guide the cursor freely in any direction.

### 3.1.1 Experimental Phases

The CP dataset consists of two experimental phases:

*1. Phase I – Subject-Specific Decoding.*

In the first phase, each participant completed eight sessions of real-time EEG-based motion control. After each session, subject-specific models were trained using the participant's cumulative data. This setup corresponds to a rehearsal-based (or memory-replay) strategy when all previous data are mixed, or cumulative learning when earlier sessions are retained intact.

In practice, although this strategy maximizes performance by exploiting the full dataset, it becomes computationally impractical as data volume increases indefinitely.

*2. Phase II – Cross-Subject Generalization and Online Adaptation.*

In the second phase, a new cohort of subjects was recruited to evaluate generalized deep-learning models. A base model trained on data from all participants in Phase I was fine-tuned for each new subject using their own data after every session. This procedure represents a transfer-learning (TL) or subject-adaptation strategy.

Because our study focuses on online adaptation performance, we used data from Phase II.

### 3.1.2 Experimental Design

During Phase II, each subject performed four sessions, each comprising 12 runs (also referred to as blocks). Each run contained five trials, with each trial lasting approximately 60 seconds. Three adaptation methods were implemented:

- AR (Autoregressive model) – traditional feature-based baseline (used only in the first session);
- DL (Deep Learning) – EEGNet trained separately for each session;
- TL (Transfer Learning) – EEGNet fine-tuned cumulatively across sessions;
- CL (Calibrated Learning) – real-time adaptation using a pre-trained baseline EEGNet decoder.

Each adaptation strategy was tested in four runs per session (the first session contained AR instead of DL). Thus, every subject contributed roughly four hours of EEG recordings, and the analysis in this study included four subjects, totaling approximately 16 hours of continuously recorded EEG data.

### 3.1.3 Dataset Structure and Exploratory Analysis

The EEG signals were recorded using 64-channel Neuroscan Quik-Caps, following the international 10–20 electrode placement system. The raw EEG data were sampled at 1000 Hz and subsequently band-pass filtered between 0.1 Hz and 60 Hz to remove slow drifts and high-frequency noise. The filtered signals were then transmitted to the Neuroscan acquisition software in blocks at 25 Hz for real-time processing, analysis, and storage.

Each recording file contains continuous EEG signals (62 active channels), accompanied by synchronized cursor kinematics (position and velocity) and target trajectories across multiple runs within each session. The data are stored in MATLAB .mat format, with filenames encoding metadata such as subject ID, session number, condition, and run index (e.g., S16_Se02_CL_R01). We organized these recordings into structured Python dictionaries using HDF5 and NumPy arrays to facilitate efficient data loading and indexing.

To gain an initial understanding of the dataset, we conducted exploratory analyses by visualizing behavioral time series and cursor–target trajectories from representative runs. Figures 1–3 illustrate sample trajectories and temporal alignments between the cursor and moving target. For example, in Subject S15, Run 1, the cursor position time series contains 7,931 samples, which, at 25 Hz, corresponds to approximately 317 seconds (~5 minutes), consistent with the duration of one continuous trial block.

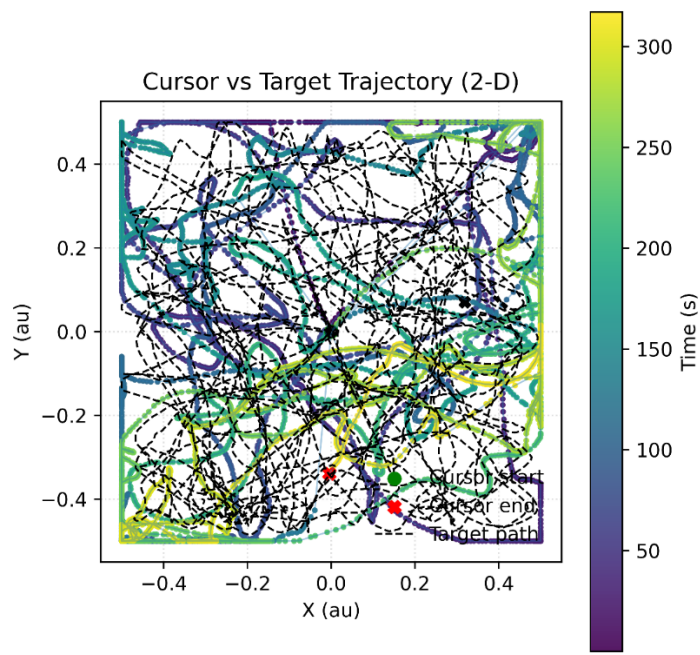

Figure 1. A representative run from the CP dataset showing the cursor path and the target path in the X–Y plane (arbitrary units). Markers indicate start (green) and end (red) locations.

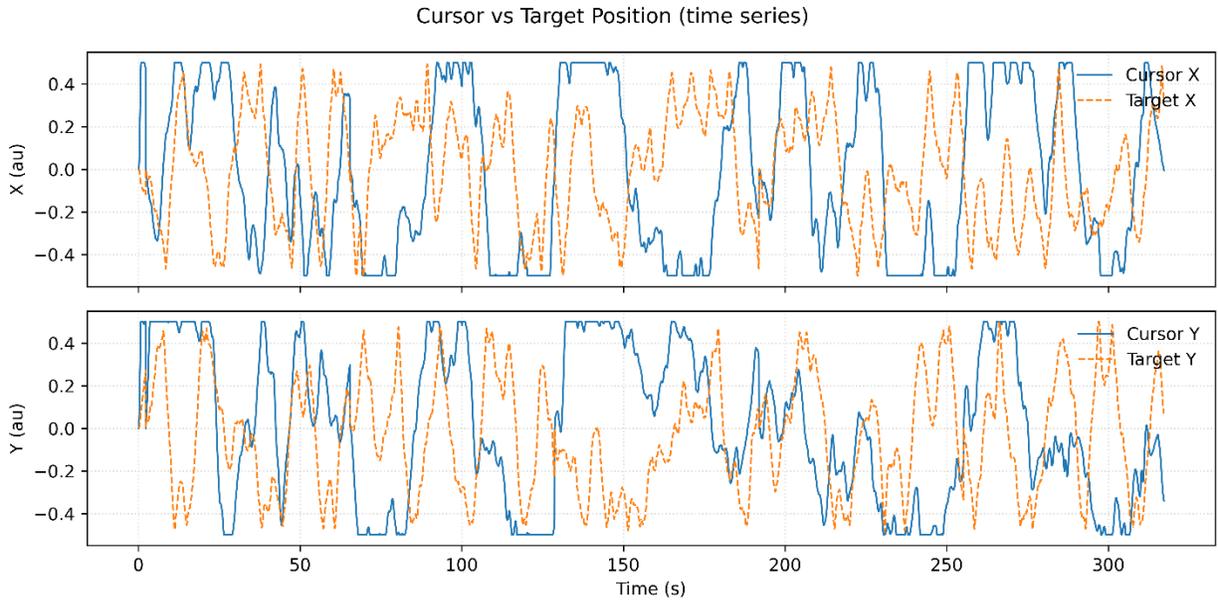

Figure 2. Temporal traces of cursor and target positions along the X (top) and Y (bottom) axes for the same run as Figure 1.

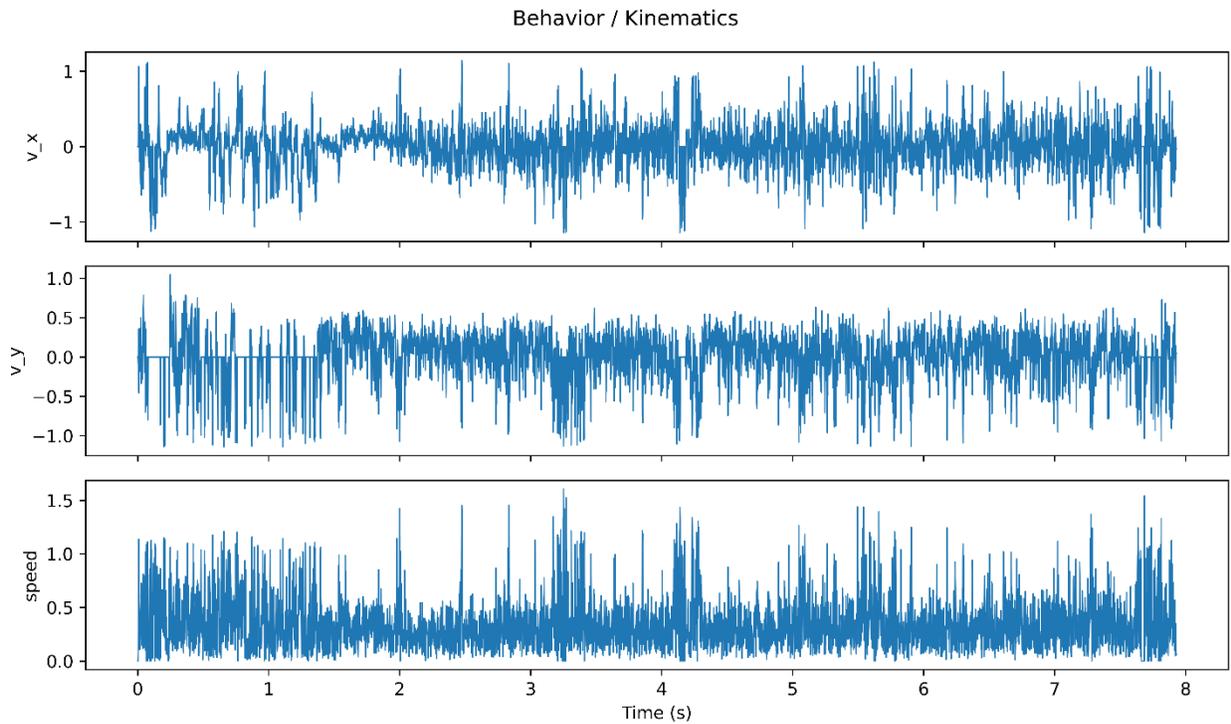

Figure 3. Horizontal velocity ($v_x$), vertical velocity ($v_y$), and cursor speed from a representative time period.

For EEG preprocessing and quality inspection, we imported the signals into MNE-Python for further analysis. To evaluate signal characteristics, we computed the power spectral density (PSD) across all channels. As shown in Figure 4, the channel-wise PSD exhibits the expected 1/f spectral profile and a notch around 60 Hz, reflecting suppression of line-frequency interference.

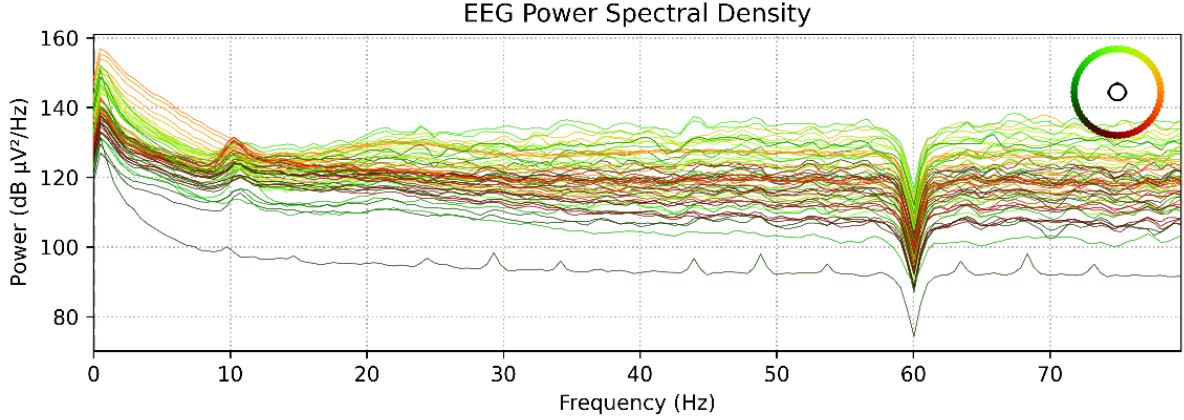

Figure 4. EEG power spectral density across channels for a representative session.

### 3.2 Preprocessing

#### 3.2.1 EEG Preprocessing and Packetization of Features

Raw EEG was downsampled from 1000 Hz to 250 Hz ($f_s$) using zero-phase decimation. Thereafter, EEG signals were converted into packets to match the control update rate ($f_p = 25 Hz$). Packets were defined by a sliding window of length

$$L = round(window\_sec \cdot f_s) \quad (1)$$

advanced every

$$H = round(step\_sec \cdot f_s) \quad (2)$$

where we adopted $window_{sec} = 0.25s$, and $step_{sec} = 0.04s$. This yields an effective packet interval of $\Delta t = 0.04s$.

For each packet, PSD was estimated with Welch's method (50% overlap, $n_{perserg} = \min(L, 256)$) on each channel. Bandpower was then computed for predefined frequency ranges $\mathcal{B}_k$ (θ [4–7 Hz], α [8–13 Hz], β [13–30 Hz]):

$$P_{c,k} = \int_{f \in \mathcal{B}_k} PSD_c(f) df \quad (3)$$

and concatenated across channels and bands to form one feature vector per packet. The resulting feature matrix is

$$X_{bp} \in \mathbb{R}^{N \times D}, D = C \times \#bands \quad (4)$$

These bandpower features are used for AR and Bayes decoders.

For CNN input, packet windows were processed with Common average reference (CAR). At each time point $t$, the average across channels was subtracted:

$$\tilde{x}_{t,c} = x_{t,c} - \frac{1}{C}\sum_{c'=1}^{C} x_{t,c'} \quad (5)$$

Next, exponential moving standardization was applied. For each channel $c$, recursive estimates of the mean and variance were maintained:

$$m_t = (1-\alpha)m_{t-1} + \alpha x_{t,c} \quad (6)$$

$$v_t = (1-\alpha)v_{t-1} + \alpha(x_{t,c} - m_t)^2 \quad (7)$$

and samples were standardized as

$$\tilde{x}_{t,c} = \frac{x_{t,c} - m_t}{\sqrt{v_t + \varepsilon}} \quad (8)$$

with smoothing factor $\alpha$ and numerical stabilizer $\varepsilon$; windows of length $L$ were extracted with hop $H$, reshaped as

$$X_{raw} \in \mathbb{R}^{N \times 1 \times C \times L} \quad (9)$$

which served as the input to EEGNet.

### 3.2.2 Label Construction and Resampling

Decoder targets were constructed according to dataset availability. If both cursor and target positions were present:

$$y_t = target_t - cursor_t \quad (10)$$

representing the position error vector. Otherwise, cursor velocities were used:

$$y_t = v_t \quad (11)$$

All label signals were linearly resampled to the packet timestamps $\{t_k\}$ to align exactly with feature packets:

$$Y[k] = interp(y(t), t, t_k), \quad Y \in \mathbb{R}^{N \times 2} \quad (12)$$

Subsequently, since bandpower and raw window packetizes can produce slightly different counts (due to edge effects or adaptive window shortening), the feature and label streams were truncated to the common length

$$N = \min(N_{bp}, N_{raw}, N_Y) \quad (13)$$

Thus, the final aligned dataset for each run consisted of $X_{bp} \in \mathbb{R}^{N \times D}$ (bandpower features), $X_{raw} \in \mathbb{R}^{N \times 1 \times C \times L}$ (raw windows), and $Y \in \mathbb{R}^{N \times 2}$ (labels, velocity or acceleration).

## 3.3 Models

### 3.3.1 Bayes Decoder

We design a Bayesian regression decoder to map neural features $X_t \in \mathbb{R}^D$ to kinematic variables $y_t \in \mathbb{R}^D$. At each time step $t$, the generative model is defined as:

$$y_t = X_t W + \varepsilon_t, \qquad \varepsilon_t \sim \mathcal{N}(0, \sigma^2 I_2) \tag{14}$$

where $W \in \mathbb{R}^{D \times 2}$ denotes the weight matric and $\sigma^2$ the observation noise variance.

We impose a Gaussian prior on the weights:

$$p(W) = \mathcal{N}(0, A^{-1}) \tag{15}$$

with precision matrix $A$. Two formulations of $A$ are considered:

1) Isotropic prior: $A = \alpha I_D$, which enforces uniform shrinkage on all features.
2) Automatic Relevance Determination(ARD) prior: $A = diag(\alpha_1, \ldots, \alpha_D)$, where irrelevant features acquire large $\alpha_j$ to effectively prune them.

Given calibration data $(X, Y)$, with sufficient statistics

$$S_{xx} = X^\top X, \quad S_{xy} = X^\top Y \tag{16}$$

the posterior mean of the weights is obtained by solving:

$$\widehat{W} = (S_{xx} + \sigma^2 A)^{-1} S_{xy} \tag{17}$$

This yields a closed-form Bayesian ridge regression solution regularized by the prior precision.

The decoder is trained using all data from previous sessions as calibration data. The sufficient statistics $(S_{xx}, S_{xy})$ are accumulated, and weights $\widehat{W}$ are computed as above. In mid-session update, the first half of the session is used to train the decoder.

The updated decoder is evaluated on the remaining half of the session. Because of the online adaption setup, the decoder continues to adapt during evaluation. At every time step, predictions are generated as

$$\hat{y}_t = X_t \widehat{W}_{t-1} \tag{18}$$

Residuals are computed:

$$r_t = y_t - \hat{y}_t \tag{19}$$

Every $K$ samples (update interval), the sufficient statistics are updated:

$$S_{xx} \leftarrow \lambda S_{xx} + X_b^\top X_b, \quad S_{xy} \leftarrow \lambda S_{xy} + X_b^\top Y_b \tag{20}$$

where $X_b, Y_b$ denote buffered mini-batches, and $\lambda \in (0,1]$ is a forgetting factor that down-weights older data. The weights are then recomputed:

$$\widehat{W} \leftarrow (S_{xx} + \sigma^2 A)^{-1} S_{xy} \tag{21}$$

After each online update, a small empirical-Bayes (EB) step is performed to gently adjust $\sigma^2$ and (for ARD) the feature-wise precision $\alpha_j$

1. Noise variance update (Exponential Weighted Moving Average on residuals):
   For residuals $r = Y - X\widehat{W}$,

$$\sigma^2 \leftarrow 0.9\sigma^2 + 0.1\,mean(r^2) \tag{22}$$

2. Approximate ARD precision update (diagonal posterior covariance approximation):
   With $\Sigma \approx (S_{xx} + \sigma^2 A)^\dagger$ and $\Sigma_{jj}$ its diagonal,

$$\gamma_j = 1 - \alpha_j \Sigma_{jj} \tag{23}$$

$$\overline{\omega}_j^2 = \frac{1}{2}\sum_{k=1}^{2} W_{jk}^2 \tag{24}$$

$$\alpha_j \leftarrow 0.9\alpha_j + 0.1\frac{\gamma_j}{\overline{\omega}_j^2} \tag{25}$$

For isotropic, a scalar $\alpha$ uses the mean $\gamma$ and mean $W^2$.

In addition, the decoder maintains a diagonal observation noise estimate $R_t$ that is updated from instantaneous residuals:

$$R_t = (1-\beta)R_{t-1} + \beta \cdot clip(r_t^{\odot 2}, R_{min}, R_{max}) \tag{26}$$

where $\beta$ is a smoothing factor. This allows the decoder to track dynamic changes in EEG noise levels.

Parameter details about the Bayesian decoder configuration is provided in Table 1.

Table 1: Bayesian decoder configuration.

| Parameter | Value | Parameter | Value |
| --- | --- | --- | --- |
| $\sigma^2$ (init) | 0.01 | Update interval | 50 packets |
| Forgetting factor | 0.98 | $\beta$ (adapt R) | 0.05 |
| $R_{min}$ | 0.001 | $R_{max}$ | 1.0 |

To account for run-to-run variability, we implemented a mid-run fine-tuning scheme for the baseline models: The first half of the run was used to estimate the model parameters $\widehat{W}$, and predictions were generated on the second half using the fixed $\widehat{W}$. This procedure closely mimics the calibration phase commonly used in practice, during which the subject provides labeled data at the beginning of a run to adapt the model before real-time control.

### 3.3.2 Autoregression (AR)

The AR baseline follows the original CP paper. The purpose is to map bandpower features $X \in \mathbb{R}^{N \times D}$ to the paired kinematic targets $Y \in \mathbb{R}^{N \times 2}$. We first standardize features component-wise:

$$\mu = \frac{1}{N}\sum_{i=1}^{N} X_i \tag{27}$$

$$\sigma = \sqrt{\frac{1}{N}\sum_{i=1}^{N}(X_i - \mu)^2 + \varepsilon}, \varepsilon = 10^{-6} \tag{28}$$

$$\widetilde{X_i} = \frac{X_i - \mu}{\sigma} \tag{29}$$

A bias term is concatenated to form the augmented design

$$X_b = [\widetilde{X}\ 1] \in \mathbb{R}^{N \times (D+1)} \tag{30}$$

We then estimate a multi-output linear map $W \in \mathbb{R}^{(D+1) \times 2}$ by ridge regression:

$$\widehat{W} = \underset{W}{\mathrm{argmin}}\ \|Y - X_b W\|_F^2 + \lambda \|W\|_F^2 \tag{31}$$

with $\lambda > 0$ the $L_2$ regularization strength. The resulting closed-form solution is

$$\widehat{W} = (X_b^\top X_b + \lambda I_{D+1})^{-1} X_b^\top Y \tag{32}$$

with regularization parameter $\lambda = 10^{-3}$.

At test time (the second half of the run), each incoming feature vector $x_{te}$ is standardized using the stored $\mu, \sigma$, augmented with a bias term, and mapped to the 2-D output by

$$\hat{y} = [\frac{x_{te} - \mu}{\sigma}\ 1]\widehat{W} \tag{33}$$

The ridge objective above is equivalent to a maximum a posteriori (MAP) solution under a Gaussian observation model with isotropic residual covariance (i.e., the two output dimensions are jointly penalized via squared error without per-axis reweighting) and a zero-mean Gaussian prior on $W$ with spherical covariance $\propto \lambda^{-1}I$. In our implementation, $\lambda$ is a direct regularize; we do not estimate a noise variance parameter, nor do we introduce output-axis-specific noise weights.

### 3.3.3 EEGNet

EEGNet was used as a baseline given its broad adoption in most latest BCI online control experiment (Lee et al., 2025).

The network follows an EEGNetv4-inspired design adapted for regression:

1. Temporal convolution (spectral filtering)
   A 2D convolution with kernel size $(1, L)$ (where $L$ is the temporal kernel length) learns temporal filters for each channel:
   
   $$H^{(1)} = BN\left(Conv_{time}(\widetilde{X})\right) \tag{34}$$
   
   with $F_1$ temporal filters.

2. Depthwise spatial convolution (channel mixing)
   A depthwise convolution across the $C$ channels captures spatial patterns:
   
   $$H^{(2)} = ELU\left(BN\left(Conv_{spatial}(H^{(1)})\right)\right) \tag{35}$$
   
   producing $F_1 \cdot D$ feature maps.

3. Pooling and dropout (regularization)
   Average pooling across channels/time and dropout are applied to reduce dimensionality and prevent overfitting.
4. Separable convolution (temporal refinement)
   A depthwise-pointwise convolution pair further refines temporal features while controlling parameter count:
$$H^{(3)} = ELU\left(BN\left(Conv_{sep}(H^{(2)})\right)\right) \tag{36}$$
   yielding $F_2$ feature maps.
5. Global pooling and dense regression head
   Adaptive average pooling collapses the spatial and temporal dimensions, and the ouput is flattened to a vector of length $F_2$. A fully connected layer maps to the final 2D output:
$$\hat{y} = W_{fc}^\top H^{(3)}_{pooled} + b_{fc}, \qquad \hat{y} \in \mathbb{R}^2 \tag{37}$$

During training, mean squired error loss was used:
$$\mathcal{L}(\theta) = \frac{1}{N}\sum_{i=1}^{N}\|y_i - f_\theta(X_i)\|_2^2 \tag{38}$$

where $f_\theta$ denotes the EEGNetCP mapping parameterized by weights $\theta$, and $(X_i, y_i)$ are the calibration samples. Optimization was performed with stochastic gradient descent.

During evaluation where the second half of the run was used, the trained model is applied to unseen windows $X^{te}$ with fixed weights $\hat{\theta}$, yielding predictions
$$\hat{y}^{te} = f_{\hat{\theta}}(X^{te}) \tag{39}$$

Details about the EEGNet architecture and training parameters are summarized in Table 2:

Table 2: EEGNetCP configuration and training parameters.

| Parameter | Value | Parameter | Value |
|---|---|---|---|
| $F_1$ | 8 | Kernel length | 64 |
| Expansion $D$ | 2 | $F_2$ | 16 |
| Pool1 size | 4 | Pool2 size | 8 (adaptive) |
| Dropout rate | 0.25 | Optimizer | Adam |
| Learning rate | 0.001 | Weight decay | 0.0 |
| Batch size | 128 | Epochs | 20 |

### 3.4 Prediction Mode

The three types of decoders were trained and evaluated in two alternative label modes, which differ in how the kinematic targets are derived from recorded trajectories.

1. Velocity mode
   The decoder is trained to map directly to the contemporaneous velocity:
   $$y_t = \begin{bmatrix} v_x(t) \\ v_y(t) \end{bmatrix} \tag{40}$$

2. Acceleration mode
   The regression targets are the finite-difference accelerations, computed as discrete changes in velocity:
   $$a_t = \frac{v_t - v_{t-1}}{\Delta t}, v_t \in \mathbb{R}^2 \tag{41}$$

   In this mode, the decoder predicts accelerations:
   $$y_t = \begin{bmatrix} a_x(t) \\ a_y(t) \end{bmatrix} \tag{42}$$

   Therefore, the decoder's outputs correspond to instantaneous changes in motion, rather than absolute velocity. Decoded acceleration $\hat{a}_t$ is then integrated to reconstruct velocity:
   $$\hat{v}_t = \hat{v}_{t-1} + \hat{a}_t \Delta t \tag{43}$$

## 3.5 Evaluation Metric

To assess model performance, we adopted the evaluation metrics defined in the original CP dataset study (Forenzo et al., 2024a). The primary quantitative measure is the Normalized Mean Squared Error (NMSE) between the predicted and true cursor velocities. NMSE provides a normalized index of reconstruction accuracy that accounts for differences in movement scale across runs and subjects.

Formally, NMSE is defined as:
$$NMSE = \frac{\sum_t \|v_t - \hat{v}_t\|^2}{\sum_t \|v_t\|^2} \tag{44}$$

where $v_t$ and $\hat{v}_t$ denote the true and predicted cursor velocities at time $t$, respectively. The numerator quantifies the squared Euclidean error between predicted and actual velocity vectors, while the denominator normalizes this error by the total motion energy across the trial.

This metric directly reflects the quality of continuous control, with lower NMSE values indicating more accurate and stable trajectory tracking between the decoded EEG signals and the intended motion.

## 4 Results

To compare the performance, for each model, we have 3 sessions × 3 conditions × 4 runs × 4 subjects = 144 data points. A Friedman test comparing NMSE in the acceleration mode across models (AR, Bayes, EEGNet; see Figure 5) was significant, $\chi^2(2) = 8.00$, $p = .018$, Kendall's W = 1.00 (very large effect). A mixed-effects model (with subject as a random intercept and all 36 observations per subject) indicated that Bayes yielded significantly lower NMSE compared to both AR and EEGNet (Holm-adjusted $p < 0.001$), whereas AR and EEGNet did not differ (see Table 3). To stabilize skewness, effects were reported on the log scale and back-transformed to produce ratios of geometric means. We found Bayes's typical NMSE was ~72% lower than that of AR and EEGNet in the acceleration mode.

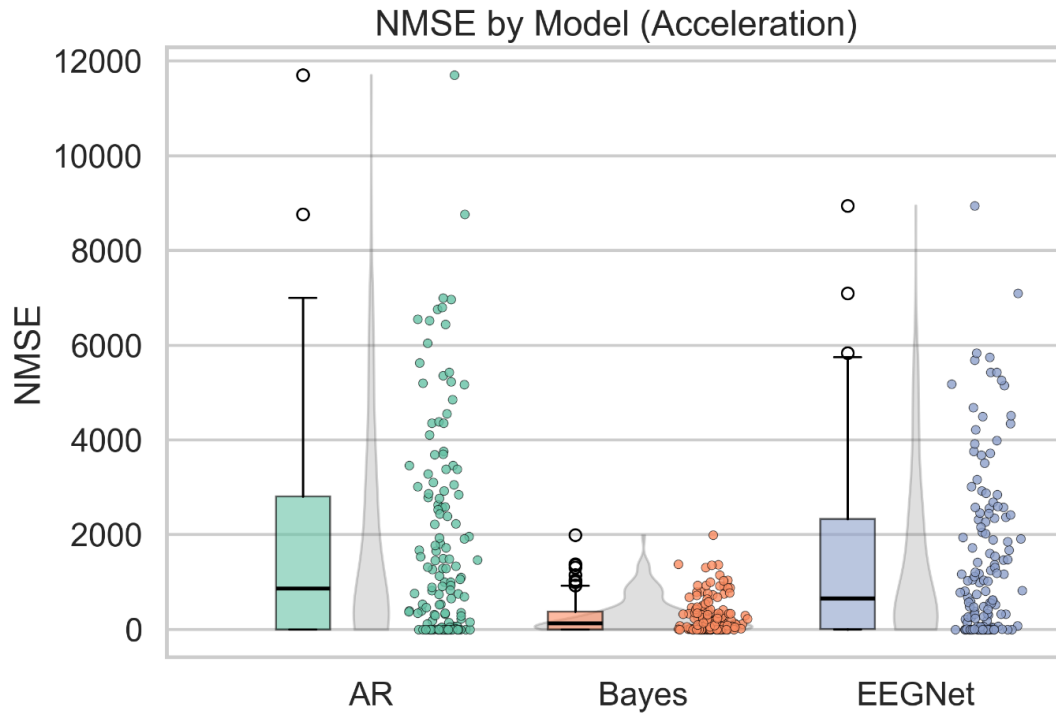

Figure 5: Box/violin/scatter plots show NMSE medians and variability by model in the acceleration mode.

Table 3 Pairwise comparisons (acceleration mode).

| Contrast | log scale est | Ratio geom means | p_uncorrected | p_holm | Sig |
| --- | --- | --- | --- | --- | --- |
| AR – Bayes | +1.29 | **3.62× higher** | 1.2e-4 | 3.5e-4 | ✓ |
| AR – EEGNet | -0.0017 | ≈ 1.00 | 0.996 | 0.996 | ✗ |
| Bayes – EEGNet | -1.29 | **0.28× (72% lower)** | 1.2e-4 | 3.5e-4 | ✓ |

However, in the velocity mode no significant differences between models were observed (all Holm-adjusted p = 1.0; see Figure 6 and Table 4).

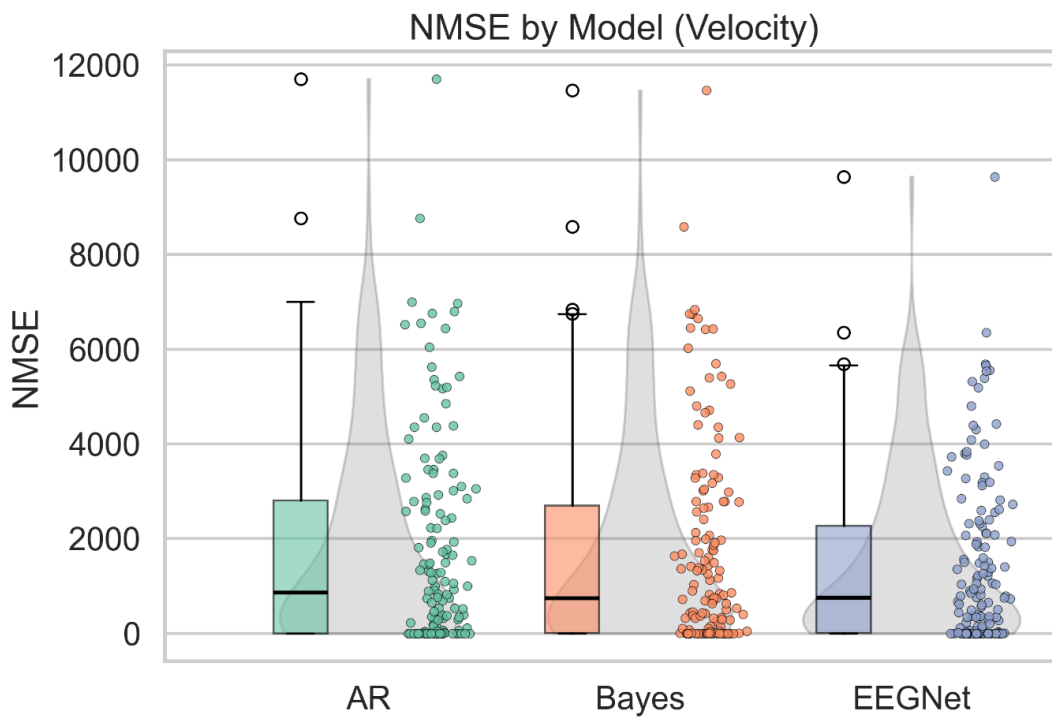

Figure 6: Box/violin/scatter plots show NMSE medians and variability by model in the velocity mode.

Table 4 Pairwise comparisons (velocity mode).

| Contrast | log scale est | Ratio geom means | p_uncorrected | p_holm | Significant? |
|---|---|---|---|---|---|
| AR – Bayes | -0.150 | 0.86× | 0.672 | 1.000 | ✗ |
| AR – EEGNet | -0.018 | 0.98× | 0.959 | 1.000 | ✗ |
| Bayes – EEGNet | +0.131 | 1.14× | 0.711 | 1.000 | ✗ |

Figure 7 illustrates NMSE trajectories across 4 runs per session (12 runs total) for each subject (S16–S19) in the acceleration mode. Across all subjects, Bayes consistently yielded lower NMSE values and remained more stable across runs (subject-level medians ~112–175, SD ≈ 275–517). In contrast, AR produced substantially higher NMSE with greater variability (medians ~700–1177, SD ≈ 1870–2500), whereas EEGNet showed intermediate performance (medians ~525–1037, SD ≈ 1500–2008). Between the two higher-error models, EEGNet exhibited lower NMSE and smaller variability compared to AR, a pattern consistent with the results reported in original study (**xx**).

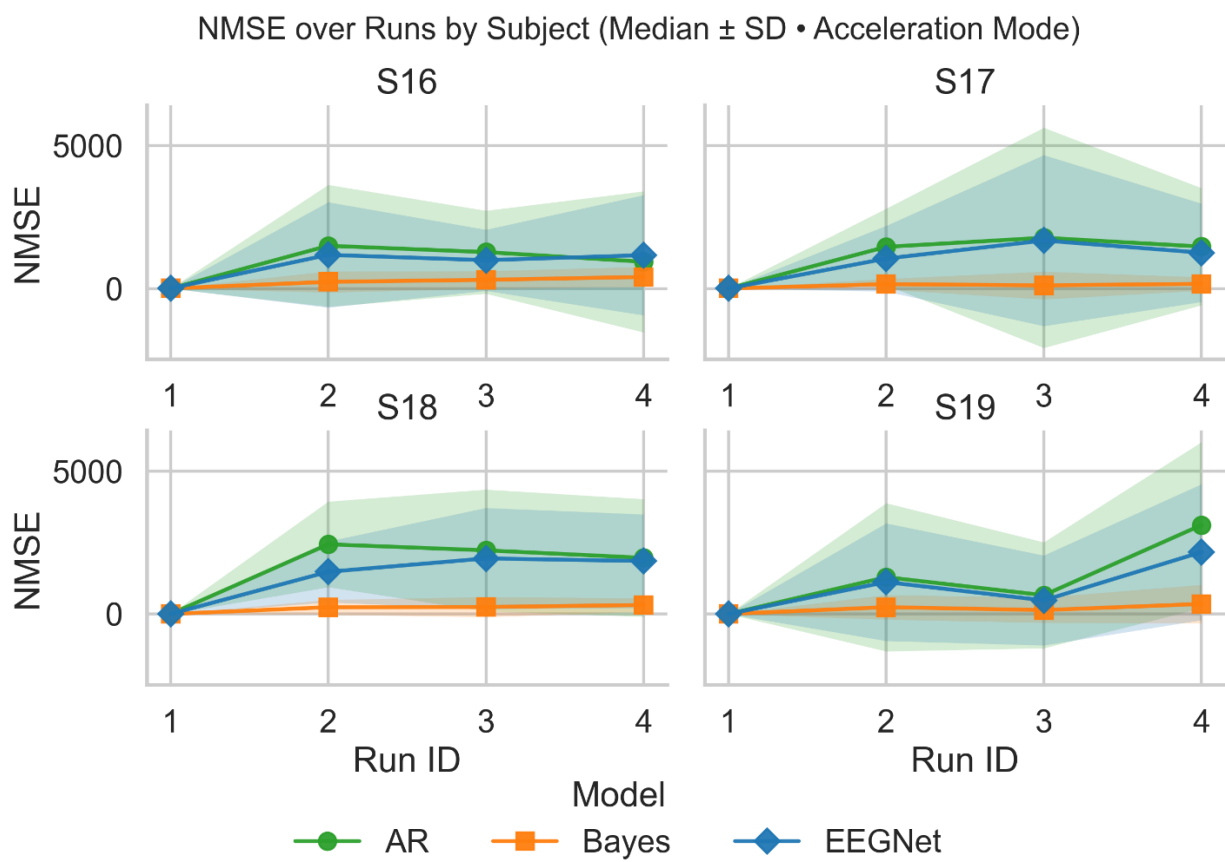

Figure 7. NMSE over runs by subject in the acceleration mode. Each panel shows a subject (S16–S19), with lines representing the median NMSE across trials per run and shaded areas indicating ±1 SD.

We also compared ARD and isotropic Bayesian models under both acceleration and velocity decoding modes on a subject-wise basis. For Subject 16 in the acceleration mode, NMSE was significantly lower under the ARD condition than under the isotropic condition, as indicated by a Wilcoxon signed-rank test (n = 39 paired observations, including one UNK condition per session from the original CP dataset), W = 104.0, $p = 2.20 \times 10^{-5}$, effect size r = 0.68 (large effect) (see Figure 8). Per-session and per-condition Wilcoxon tests (Holm-corrected within family) showed that these differences arose primarily from Sessions 2 and 3 and the DL condition (Tables 5–6).

In contrast, no significant difference was observed between the ARD and isotropic Bayesian models in the velocity mode for Subject 16 (W = 377.0, p = 0.863, r ≈ 0.03; Figure 9). These results suggest that the superior performance of the Bayesian model in the acceleration mode is not random but reflects its ability to capture feature relevance and functional importance more effectively. This advantage is particularly meaningful for intermediate BCI sessions, where users are still stabilizing their motor imagery strategies and the decoder operates without explicit adaptation mechanisms.

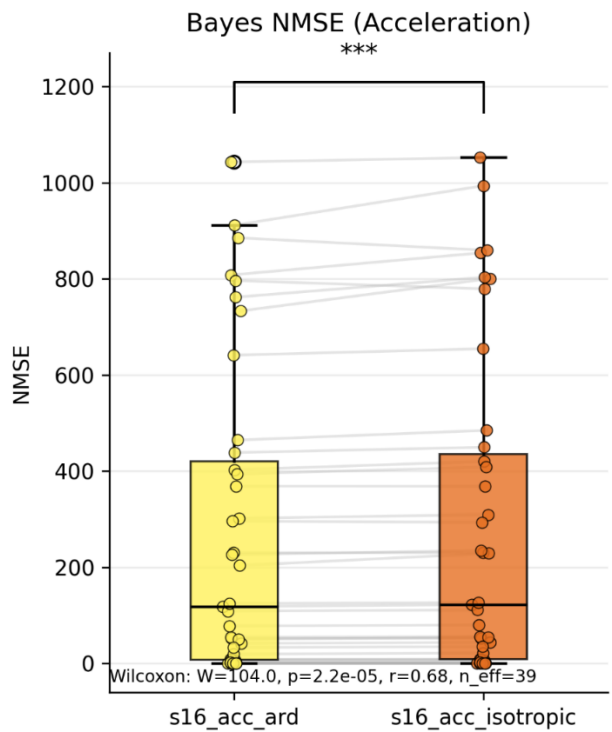

Figure 8. Bayes NMSE comparisons between ARD and isotropic setups for Subject 16 in the acceleration mode.

Table 6. Per-session Wilcoxon comparisons for Subject 16 in the acceleration mode (ARD vs isotropic Bayes NMSE).

| Session | Median ard | Median isotropic | Median Δ (isotropic −ard) | Mean Δ | W | p_uncor | p_holm | Significant? |
|---|---|---|---|---|---|---|---|---|
| Se02 | 230.5 | 230.2 | +3.74 | +18.5 | 3 | 0.0012 | **0.0037** | ✓ |
| Se03 | 41.7 | 44.1 | +0.35 | +4.0 | 12 | 0.0171 | **0.0034** | ✓ |
| Se04 | 226.5 | 234.9 | +2.10 | +3.8 | 23 | 0.1272 | 0.1272 | ✗ |

Table 7. Per-condition Wilcoxon comparisons for Subject 16 in the acceleration mode (ARD vs isotropic Bayes NMSE).

| Condition | Median ard | Median isotropic | Median Δ | Mean Δ | W | p_uncor | p_holm | Significant? |
|---|---|---|---|---|---|---|---|---|
| CL | 93.4 | 95.7 | +0.40 | +2.4 | 14 | 0.0522 | 0.1567 | ✗ |
| DL | 48.2 | 50.0 | +2.38 | +13.7 | 0 | 0.00049 | **0.00195** | ✓ |
| TL | 385.8 | 394.4 | +3.79 | +12.5 | 14 | 0.0522 | 0.1567 | ✗ |
| UNK | 124.6 | 126.7 | +0.45 | +0.1 | 3 | 1.0000 | 1.0000 | ✗ |

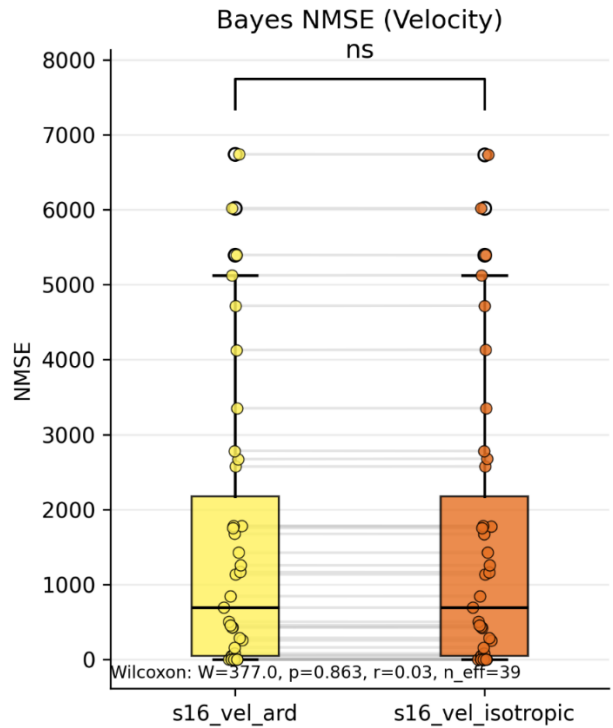

Figure 9. Bayes NMSE comparisons between ARD and isotropic setups for Subject 16 in the velocity mode.

## 5 Discussion

Our findings suggest that motor intentions in the human brain are expressed more naturally in acceleration space than in velocity space. Although position is the observable outcome and acceleration is a second-order derivative, decoding EEG activity at the acceleration level produces trajectories that are more closely aligned with the brain's natural control logic.

### 5.1 Scientific Significance: Evidence for Embodied Dynamics

At the cognitive and neurophysiological levels, these results provides empirical support for longstanding debates in neuroscience and cognitive science regarding the representational basis of motor control (Omrani et al., 2017). The evidence challenges the view that the brain encodes abstract kinematic parameters such as spatial position (Scott, 2008), instead supporting the perspective that motor control is organized through dynamical variables, specifically, force and acceleration (Michaels et al., 2016; Pandarinath et al., 2015).

Within the framework of embodied cognition, this finding implies that cognition emerges from the continuous regulation of sensorimotor dynamics instead of detached positional codes. Acceleration holds a privileged position in this control hierarchy: it can be directly sensed by the vestibular system that responds to inertial and gravitational forces (Green et al., 2005), as well as by proprioceptive feedback from muscles and joints (Proske & Gandevia, 2012). Neurophysiological studies further corroborate this view: neurons in the primary motor cortex show stronger correlations with acceleration and applied force than with absolute position (Kalaska et al., 1990), while cerebellar and basal ganglia circuits also specialize in predictive error correction at the level of dynamic forces and accelerations (Haar & Donchin, 2020). Moreover, these findings highlight a central principle that cognition is embedded not in the discrete end-point specification but in the ongoing regulation of process variables (Hatsopoulos et al., 2007).

## 5.2 Implications for Digital Twin of Robotics and Embodied AI

A particular striking observation is that this embodied preference for acceleration persisted even in a virtual, MI-based BCI task. Participants controlled a cursor purely through imagined movement without actual limb movement, muscle force output, or vestibular feedback. Yet, their neural activity still reflected acceleration-based control. This finding suggests that embodiment is not confined to overt physical interaction but extends to internal simulations of the body's sensorimotor apparatus. In other words, embodiment can manifest in the absence of physical sensors and actuators or external movement, as long as internal representations of the body's dynamic structure remain engaged. This interpretation is consistent with neuroimaging evidence showing overlapping neural substrates for motor imagery and execution (Behrendt et al., 2021; Boulenger et al., 2006; Sobierajewicz et al., 2017; Sobierajewicz et al., 2016) and with theories of internal simulation of action (Jeannerod, 2001).

These findings carry important implications for embodied robotics and artificial intelligence (AI). For example, a robotic system that internally encodes and regulates motion in terms of force and acceleration, analogous to neural representations in biological systems, may thus achieve functional embodiment even in purely virtual environments. Contemporary motor control architectures for avatars or simulated agents are typically organized around position trajectories or velocity profiles (Lanillos et al., 2021), whereas decades of robotics research show the benefits of force/impedance and operational-space dynamics (Hogan, 1984; Khatib & Burdick, 1987). Our findings suggest that acceleration- and force-based representations, though long recognized in physical control, should also underpin virtual and cognitive forms of embodiment.

By grounding EEG decoding in the same dynamic principles that govern biological motion, a direct benefit for BCI research is that brain activity decoding can more accurately reflect the brain's intrinsic representation of action, thereby enhancing control stability and reducing calibration demands. In the longer term, this approach can foster more natural and trustworthy human–robot interactions, improve collaboration performance, and broadly contribute to bridging and complementing embodied robotic intelligence with human neural processes.

## 5.3 Limitations and Future Work

Despite these promising results, several limitations warrant considerations.

First, this study relied on a publicly available dataset rather than newly collected experimental data. While benchmark datasets provide a controlled and reproducible testbed widely recognized in the BCI community, offline analyses inherently lack the proprioceptive and vestibular feedback that participants experience during real-time control. Nevertheless, the proposed Bayesian framework achieved the best performance across all conditions represented in the dataset, highlighting its robustness and potential for real-world application. As an immediate next step, we plan to conduct real-time user studies using a robotic wheelchair platform (Xu et al., 2025), enabling participants to perform continuous driving tasks in naturalistic indoor environments and test the model's ability to adapt online and support embodied interaction .

Second, when comparing Bayesian inference with deep-learning approaches, we employed EEGNet as a representative neural network baseline. Although more recent architectures, such as NeuroGPT (Cui et al., 2024) and EEGConformer (Song et al., 2022)) have demonstrated improved performance, they still operate under the same transfer learning paradigm considered here. Moreover, their greater computational complexity poses challenges for real-time deployment. In our experiments, EEGNet training required approximately five times longer than Bayesian adaptation, indicating not only the efficiency advantage of probabilistic inference but also its suitability for on-device and adaptive BCI control where latency is critical.

Third, the current work primarily emphasized decoding accuracy, while post-decoding trajectory regulation and control smoothness remain open areas for improvement. Future studies should explore how hierarchical regulatory mechanisms can be integrated to enhance stability without compromising responsiveness. In particular, mechanisms grounded in kinematic invariants such as the minimum-jerk principle (Todorov, 2004), may help regulate trajectories in the CP task. Furthermore, although our current framework tracks residual noise term $R_t$, this term was not yet incorporated into the MAP update. In future extensions, especially under shared-control settings or when fused with Kalman filtering, explicitly incorporating $R_t$ as a measurement noise term could enable more principled modeling of uncertainty and further enhance closed-loop adaptability.

## 6 Conclusion

This study introduced a Bayesian inference framework for decoding EEG signals in CP motion control, exploring an alternative hypothesis that the human brain organizes motor intentions more naturally in acceleration space than in velocity or positional space. Through analysis of the CP BCI dataset, we demonstrated that acceleration-level decoding not only yields superior performance (reduced NMSE by 72%) but also aligns with the brain's intrinsic control dynamics, reflecting its embodied and predictive nature.

From a scientific standpoint, these findings contribute empirical evidence to theories of embodied cognition and motor control, supporting the view that human movement is regulated through dynamic variables such as force and acceleration rather than abstract positional representations. Even in a purely motor imagery context devoid of physical movement, neural activity preserved this acceleration-based organization, indicating that embodiment persists at the level of internal body schema and predictive simulation.

At the computational level, the results highlight the advantages of probabilistic inference over deterministic deep learning models for real-time BCI applications. The Bayesian approach proved both computationally efficient and biologically interpretable, capable of adapting to uncertainty and capturing the hierarchical structure of neural control. These properties make it particularly suited for adaptive and on-device systems.

Practically, grounding EEG decoding in the same dynamical principles that govern biological motion offers a promising path toward more stable and intuitive BCI control. In the long term, this approach may enable seamless human–robot collaboration, improve trust and transparency in shared-control systems, and inform the design of embodied AI and digital twin frameworks that bridge neural computation with physical and virtual autonomy.

Future work will extend this framework to real-time experiments with robotic wheelchair platforms, integrating acceleration-level decoding with adaptive feedback and shared-control mechanisms. Through this, we aim to bring BCI research closer to its ultimate goal, realizing embodied, adaptive, and trustworthy human–machine interaction that reflects the intelligence and flexibility of the human brain.

## Acknowledgement

The work presented in this dissertation proposal was supported financially by the United States National Science Foundation (NSF) via Award# SCC-IRG 2124857. The support of the NSF is gratefully acknowledged.

## References

AARP. (2024). About AARP Livable Communities. Retrieved from https://www.aarp.org/livable-communities/about/?irgwc=1&s3=xHFS9zR17xyKW8eW4uyiBX-WUkCT8pxgr00ETk0&lgt=true&pubval=imp&cmp=EMC-MIV-ACQ-imp-link-pros-ctrl-a-join-2024#:~:text=By%202030,%20that%20number%20will,older%20adults%20than%20of%20children.#:~:text=By%202030,%20that%20number%20will,older%20adults%20than%20of%20children.###

Abiyev, R. H., Akkaya, N., Aytac, E., Günsel, I., & Çağman, A. (2016). Brain-Computer Interface for Control of Wheelchair Using Fuzzy Neural Networks. *Biomed Res Int, 2016*, 9359868.

Al-Qaysi, Z. T., Zaidan, B. B., Zaidan, A. A., & Suzani, M. S. (2018). A review of disability EEG based wheelchair control system: Coherent taxonomy, open challenges and recommendations. *Comput Methods Programs Biomed, 164*, 221-237.

Aller, M., & Noppeney, U. (2019). To integrate or not to integrate: Temporal dynamics of hierarchical Bayesian causal inference. *PLoS Biol, 17*(4), e3000210.

Altaheri, H., Muhammad, G., Alsulaiman, M., Amin, S. U., Altuwaijri, G. A., Abdul, W., Bencherif, M. A., & Faisal, M. (2023). Deep learning techniques for classification of electroencephalogram (EEG) motor imagery (MI) signals: A review. *Neural Computing and Applications, 35*(20), 14681-14722.

Altuwaijri, G. A., Muhammad, G., Altaheri, H., & Alsulaiman, M. (2022). A multi-branch convolutional neural network with squeeze-and-excitation attention blocks for EEG-based motor imagery signals classification. *Diagnostics, 12*(4), 995.

Amin, S. U., Altaheri, H., Muhammad, G., Alsulaiman, M., & Abdul, W. (2021). *Attention based Inception model for robust EEG motor imagery classification.* Paper presented at the 2021 IEEE international instrumentation and measurement technology conference (I2MTC).

Ang, K. K., Chin, Z. Y., Wang, C., Guan, C., & Zhang, H. (2012). Filter bank common spatial pattern algorithm on BCI competition IV datasets 2a and 2b. *Frontiers in Neuroscience, 6*, 39.

Baldi, P., & Itti, L. (2010). Of bits and wows: A Bayesian theory of surprise with applications to attention. *Neural Networks, 23*(5), 649-666.

Beck, J. M., Ma, W. J., Pitkow, X., Latham, P. E., & Pouget, A. (2012). Not noisy, just wrong: the role of suboptimal inference in behavioral variability. *Neuron, 74*(1), 30-39.

Behrendt, F., Zumbrunnen, V., Brem, L., Suica, Z., Gäumann, S., Ziller, C., Gerth, U., & Schuster-Amft, C. (2021). Effect of Motor Imagery Training on Motor Learning in Children and Adolescents: A Systematic Review and Meta-Analysis. *Int J Environ Res Public Health, 18*(18).

Belkacem, A. N., Jamil, N., Palmer, J. A., Ouhbi, S., & Chen, C. (2020). Brain Computer Interfaces for Improving the Quality of Life of Older Adults and Elderly Patients. *Frontiers in Neuroscience, 14*.

Bezyak, J., Sabella, S., Hammel, J., McDonald, K., Jones, R., & Barton, D. (2019). Community participation and public transportation barriers experienced by people with disabilities. *Disability and Rehabilitation, 42*, 1-9.

Bi, L., Fan, X. A., & Liu, Y. (2013). EEG-Based Brain-Controlled Mobile Robots: A Survey. *IEEE Transactions on Human-Machine Systems, 43*(2), 161-176.

Blankertz, B., Müller, K. R., Krusienski, D. J., Schalk, G., Wolpaw, J. R., Schlögl, A., Pfurtscheller, G., Millán Jdel, R., Schröder, M., & Birbaumer, N. (2006). The BCI competition. III: Validating alternative approaches to actual BCI problems. *IEEE Trans Neural Syst Rehabil Eng, 14*(2), 153-159.

Boulenger, V., Roy, A. C., Paulignan, Y., Deprez, V., Jeannerod, M., & Nazir, T. A. (2006). Cross-talk between language processes and overt motor behavior in the first 200 msec of processing. *J Cogn Neurosci, 18*(10), 1607-1615.

Brooks, R., Hassabis, D., Bray, D., & Shashua, A. (2012). Is the brain a good model for machine intelligence? *Nature, 482*(7386), 462.

Brouwer, A.-M., Franz, V. H., & Thornton, I. M. (2004). Representational momentum in perception and grasping: Translating versus transforming objects. *Journal of Vision, 4*(7), 575-584.

Buzsáki, G. (2006). *Rhythms of the Brain*: Oxford university press.

Candrea, D. N., Shah, S., Luo, S., Angrick, M., Rabbani, Q., Coogan, C., Milsap, G. W., Nathan, K. C., Wester, B. A., Anderson, W. S., Rosenblatt, K. R., Uchil, A., Clawson, L., Maragakis, N. J., Vansteensel, M. J., Tenore, F. V., Ramsey, N. F., Fifer, M. S., & Crone, N. E. (2024). A click-based electrocorticographic brain-computer interface enables long-term high-performance switch scan spelling. *Communications Medicine, 4*(1), 207.

CDC. (2024). CDC Data Shows Over 70 Million U.S. Adults Reported Having a Disability. Retrieved from https://www.cdc.gov/media/releases/2024/s0716-Adult-disability.html#:~:text=The%20latest%20data%2C%20from%20the,having%20a%20disability%20in%202022.

Chen, W., Chen, S.-K., Liu, Y.-H., Chen, Y.-J., & Chen, C.-S. (2022). An Electric Wheelchair Manipulating System Using SSVEP-Based BCI System. *Biosensors, 12*(10), 772.

Chiappa, S. (2006). *Analysis and classification of EEG signals using probabilistic models for brain computer interfaces.* Verlag nicht ermittelbar,

Cui, W., Jeong, W., Thölke, P., Medani, T., Jerbi, K., Joshi, A. A., & Leahy, R. M. (2024). *Neuro-gpt: Towards a foundation model for eeg.* Paper presented at the 2024 IEEE International Symposium on Biomedical Imaging (ISBI).

Davies, A., De Souza, L., & Frank, A. (2003). Changes in the quality of life in severely disabled people following provision of powered indoor/outdoor chairs. *Disability and Rehabilitation, 25*, 286-290.

Dawson, M. R. W. (2019). Embodied Perception. In J. Vonk & T. Shackelford (Eds.), *Encyclopedia of Animal Cognition and Behavior* (pp. 1-8). Cham: Springer International Publishing.

Fahimi, F., Zhang, Z., Goh, W. B., Lee, T. S., Ang, K. K., & Guan, C. T. (2019). Inter-subject transfer learning with an end-to-end deep convolutional neural network for EEG-based BCI. *Journal of Neural Engineering, 16*(2).

Fernández-Rodríguez, Á., Velasco-Álvarez, F., & Ron-Angevin, R. (2016). Review of real brain-controlled wheelchairs. *J Neural Eng, 13*(6), 061001.

Fiser, J., Berkes, P., Orbán, G., & Lengyel, M. (2010). Statistically optimal perception and learning: from behavior to neural representations. *Trends in Cognitive Sciences, 14*(3), 119-130.

Fitts, P. M., & Peterson, J. R. (1964). Information capacity of discrete motor responses. *Journal of experimental psychology, 67*(2), 103.

Flash, T., & Hogan, N. (1985). The coordination of arm movements: an experimentally confirmed mathematical model. *Journal of Neuroscience, 5*(7), 1688-1703.

Forenzo, D., Zhu, H., & He, B. (2024a). A continuous pursuit dataset for online deep learning-based EEG brain-computer interface. *Scientific Data, 11*(1), 1256.

Forenzo, D., Zhu, H., Shanahan, J., Lim, J., & He, B. (2024b). Continuous tracking using deep learning-based decoding for noninvasive brain–computer interface. *PNAS Nexus, 3*(4).

Frank, A. O., Ward, J., Orwell, N. J., McCullagh, C., & Belcher, M. (2000). Introduction of a new NHS electric-powered indoor/outdoor chair (EPIOC) service: benefits, risks and implications for prescribers. *Clin Rehabil, 14*(6), 665-673.

Freitas, J., Teixeira, A., Dias, M. S., & Silva, S. (2017). *An introduction to silent speech interfaces*: Springer.

Freyd, J. J., & Finke, R. A. (1984). Facilitation of length discrimination using real and imaged context frames. *Am J Psychol, 97*(3), 323-341.

Friston, K. (2010). The free-energy principle: a unified brain theory? *Nature Reviews Neuroscience, 11*(2), 127-138.

Friston, K. (2012). The history of the future of the Bayesian brain. *Neuroimage, 62*(2), 1230-1233.

Girshick, A. R., Landy, M. S., & Simoncelli, E. P. (2011). Cardinal rules: visual orientation perception reflects knowledge of environmental statistics. *Nature Neuroscience, 14*(7), 926-932.

Gómez, C. M., Arjona, A., Donnarumma, F., Maisto, D., Rodríguez-Martínez, E. I., & Pezzulo, G. (2019). Tracking the Time Course of Bayesian Inference With Event-Related Potentials:A Study Using the Central Cue Posner Paradigm. *Frontiers in Psychology, 10*.

Green, A. M., Shaikh, A. G., & Angelaki, D. E. (2005). Sensory vestibular contributions to constructing internal models of self-motion. *J Neural Eng, 2*(3), S164-179.

Gwon, D., Won, K., Song, M., Nam, C. S., Jun, S. C., & Ahn, M. (2023). Review of public motor imagery and execution datasets in brain-computer interfaces. *Frontiers in Human Neuroscience, 17*.

Haar, S., & Donchin, O. (2020). A Revised Computational Neuroanatomy for Motor Control. *Journal of Cognitive Neuroscience, 32*(10), 1823-1836.

Hamad, E. M., Al-Gharabli, S. I., Saket, M. M., & Jubran, O. (2017, 11-15 July 2017). *A Brain Machine Interface for command based control of a wheelchair using conditioning of oscillatory brain activity.* Paper presented at the 2017 39th Annual International Conference of the IEEE Engineering in Medicine and Biology Society (EMBC).

Harris, C. M., & Wolpert, D. M. (1998). Signal-dependent noise determines motor planning. *Nature, 394*(6695), 780-784.

Hasenstab, K., Scheffler, A., Telesca, D., Sugar, C. A., Jeste, S., DiStefano, C., & Şentürk, D. (2017). A multi-dimensional functional principal components analysis of EEG data. *Biometrics, 73*(3), 999-1009.

Hassanpour, A., Moradikia, M., Adeli, H., Khayami, S. R., & Shamsinejadbabaki, P. (2019). A novel end-to-end deep learning scheme for classifying multi-class motor imagery electroencephalography signals. *Expert Systems, 36*(6), e12494.

Hatsopoulos, N. G., Xu, Q., & Amit, Y. (2007). Encoding of movement fragments in the motor cortex. *J Neurosci, 27*(19), 5105-5114.

He, S., Zhang, R., Wang, Q., Chen, Y., Yang, T., Feng, Z., Zhang, Y., Shao, M., & Li, Y. (2017). A P300-Based Threshold-Free Brain Switch and Its Application in Wheelchair Control. *Ieee Transactions on Neural Systems and Rehabilitation Engineering, 25*(6), 715-725.

Herculano-Houzel, S. (2009). The human brain in numbers: a linearly scaled-up primate brain. *Frontiers in Human Neuroscience, 3*, 857.

Hinton, G. E. (1984). Distributed representations.

Hogan, N. (1982). *Control and coordination of voluntary arm movements.* Paper presented at the 1982 American Control Conference.

Hogan, N. (1984). *Impedance control: An approach to manipulation.* Paper presented at the 1984 American control conference.

Hohwy, J. (2013). *The Predictive Mind*: Oxford University Press.

Hohwy, J. (2017). Priors in perception: Top-down modulation, Bayesian perceptual learning rate, and prediction error minimization. *Consciousness and Cognition, 47*, 75-85.
31

Hopfield, J. J. (1982). Neural networks and physical systems with emergent collective computational abilities. *Proceedings of the National Academy of Sciences, 79*(8), 2554-2558.
Hubel, D. H., & Wiesel, T. N. (1959). Receptive fields of single neurones in the cat's striate cortex. *The Journal of physiology, 148*(3), 574.
Hwang, h.-j., Lim, J.-H., Kim, D.-W., & Im, C.-H. (2014). Evaluation of various mental task combinations for near-infrared spectroscopy-based brain-computer interfaces. *Journal of biomedical optics, 19*, 77005.
Ingolfsson, T. M., Hersche, M., Wang, X. Y., Kobayashi, N., Cavigelli, L., Benini, L., & Ieee. (2020, Oct 11-14

2020). *EEG-TCNet: An Accurate Temporal Convolutional Network for Embedded Motor-Imagery Brain-Machine Interfaces.* Paper presented at the IEEE International Conference on Systems, Man, and Cybernetics (SMC), null, ELECTR NETWORK.
Inoue, Y., Mao, H., Suway, S. B., Orellana, J., & Schwartz, A. B. (2018). Decoding arm speed during reaching. *Nature Communications, 9*(1), 5243.
Iturrate, I., Antelis, J., & Minguez, J. (2009a, 12-17 May 2009). *Synchronous EEG brain-actuated wheelchair with automated navigation.* Paper presented at the 2009 IEEE International Conference on Robotics and Automation.
Iturrate, I., Antelis, J. M., Kubler, A., & Minguez, J. (2009b). A Noninvasive Brain-Actuated Wheelchair Based on a P300 Neurophysiological Protocol and Automated Navigation. *IEEE Transactions on Robotics, 25*(3), 614-627.
Jansuwan, S., Christensen, K., & Chen, A. (2013). Assessing the Transportation Needs of Low-Mobility Individuals: Case Study of a Small Urban Community in Utah. *Journal of Urban Planning and Development, 139*, 104-114.
Jeannerod, M. (2001). Neural simulation of action: a unifying mechanism for motor cognition. *Neuroimage, 14*(1 Pt 2), S103-109.
Jones, K. E., Hamilton, A. F. d. C., & Wolpert, D. M. (2002). Sources of signal-dependent noise during isometric force production. *Journal of Neurophysiology, 88*(3), 1533-1544.
Jun, S. C., George, J. S., Paré-Blagoev, J., Plis, S. M., Ranken, D. M., Schmidt, D. M., & Wood, C. C. (2005). Spatiotemporal Bayesian inference dipole analysis for MEG neuroimaging data. *Neuroimage, 28*(1), 84-98.
Kalaska, J. F., Cohen, D. A., Prud'homme, M., & Hyde, M. L. (1990). Parietal area 5 neuronal activity encodes movement kinematics, not movement dynamics. *Exp Brain Res, 80*(2), 351-364.
Khatib, O., & Burdick, J. (1987). Optimization of dynamics in manipulator design: The operational space formulation. *INT. J. ROBOTICS AUTOM., 2*(2), 90-98.
Kim, K. T., & Lee, S. W. (2016, 9-12 Oct. 2016). *Towards an EEG-based intelligent wheelchair driving system with vibro-tactile stimuli.* Paper presented at the 2016 IEEE International Conference on Systems, Man, and Cybernetics (SMC).
Knill, D. C., & Pouget, A. (2004). The Bayesian brain: the role of uncertainty in neural coding and computation. *Trends in neurosciences, 27*(12), 712-719.
Körding, K. P., & Wolpert, D. M. (2004). Bayesian integration in sensorimotor learning. *Nature, 427*(6971), 244-247.
Kotseruba, I., & Tsotsos, J. K. (2020). 40 years of cognitive architectures: core cognitive abilities and practical applications. *Artificial Intelligence Review, 53*(1), 17-94.

Krizhevsky, A., Sutskever, I., & Hinton, G. E. (2012). Imagenet classification with deep convolutional neural networks. *Advances in Neural Information Processing Systems, 25*.
Lakas, A., Kharbash, F., & Belkacem, A. N. (2021, 28 June-2 July 2021). *A Cloud-based Brain-controlled Wheelchair with Autonomous Indoor Navigation System.* Paper presented at the 2021 International Wireless Communications and Mobile Computing (IWCMC).
Lanillos, P., Meo, C., Pezzato, C., Meera, A. A., Baioumy, M., Ohata, W., Tschantz, A., Millidge, B., Wisse, M., & Buckley, C. L. (2021). Active inference in robotics and artificial agents: Survey and challenges. *arXiv preprint arXiv:2112.01871*.
Lawhern, V., Solon, A., Waytowich, N., Gordon, S., Hung, C., & Lance, B. (2016). EEGNet: A Compact Convolutional Network for EEG-based Brain-Computer Interfaces. *Journal of Neural Engineering, 15*.
LeCun, Y., Bengio, Y., & Hinton, G. (2015). Deep learning. *Nature, 521*(7553), 436-444.
Lee, J. Y., Lee, S., Mishra, A., Yan, X., McMahan, B., Gaisford, B., Kobashigawa, C., Qu, M., Xie, C., & Kao, J. C. (2025). Brain–computer interface control with artificial intelligence copilots. *Nature Machine Intelligence, 7*(9), 1510-1523.
Li, D., Xu, J., Wang, J., Fang, X., & Ji, Y. (2020). A multi-scale fusion convolutional neural network based on attention mechanism for the visualization analysis of EEG signals decoding. *Ieee Transactions on Neural Systems and Rehabilitation Engineering, 28*(12), 2615-2626.
Li, J., Liang, J., Zhao, Q., Li, J., Hong, K., & Zhang, L. (2013). Design of assistive wheelchair system directly steered by human thoughts. *Int J Neural Syst, 23*(3), 1350013.
Li, M., Feng, X., & Liu, X. (2024). 3D point-cloud data corrosion model for predictive maintenance of concrete sewers. *Automation in Construction, 168*, 105743.
Li, Z., Yu, Y., Tian, F., Chen, X., Xiahou, X., & Li, Q. (2025). Vigilance recognition for construction workers using EEG and transfer learning. *Advanced Engineering Informatics, 64*, 103052.
Liu, T., & Yang, D. (2021). A densely connected multi-branch 3D convolutional neural network for motor imagery EEG decoding. *Brain Sciences, 11*(2), 197.
Long, J., Li, Y., Wang, H., Yu, T., Pan, J., & Li, F. (2012). A Hybrid Brain Computer Interface to Control the Direction and Speed of a Simulated or Real Wheelchair. *Ieee Transactions on Neural Systems and Rehabilitation Engineering, 20*(5), 720-729.
Lopes, M., & Santos-Victor, J. (2005). Visual learning by imitation with motor representations. *IEEE Trans Syst Man Cybern B Cybern, 35*(3), 438-449.
Luis, L., & Gomez-Gil, J. (2012). Brain Computer Interfaces, a Review. *Sensors (Basel, Switzerland), 12*, 1211-1279.
Luo, T.-j., Zhou, C.-l., & Chao, F. (2018). Exploring spatial-frequency-sequential relationships for motor imagery classification with recurrent neural network. *BMC bioinformatics, 19*(1), 344.
Ma, W. J., Beck, J. M., Latham, P. E., & Pouget, A. (2006). Bayesian inference with probabilistic population codes. *Nature Neuroscience, 9*(11), 1432-1438.
Maex, R., Berends, M., & Cornelis, H. (2009). Large-scale network simulations in systems neuroscience.
Maiseli, B., Abdalla, A. T., Massawe, L. V., Mbise, M., Mkocha, K., Nassor, N. A., Ismail, M., Michael, J., & Kimambo, S. (2023). Brain-computer interface: trend, challenges, and threats. *Brain Inform, 10*(1), 20.

Michaels, J. A., Dann, B., & Scherberger, H. (2016). Neural Population Dynamics during Reaching Are Better Explained by a Dynamical System than Representational Tuning. *PLoS Comput Biol, 12*(11), e1005175.

Millan, J. d. R., Galan, F., Vanhooydonck, D., Lew, E., Philips, J., & Nuttin, M. (2009, 3-6 Sept. 2009). *Asynchronous non-invasive brain-actuated control of an intelligent wheelchair.* Paper presented at the 2009 Annual International Conference of the IEEE Engineering in Medicine and Biology Society.

Montesano, L., Lopes, M., Bernardino, A., & Santos-Victor, J. (2008). Learning Object Affordances: From Sensory--Motor Coordination to Imitation. *IEEE Transactions on Robotics, 24*(1), 15-26.

Morasso, P. (1981). Spatial control of arm movements. *Experimental Brain Research, 42*(2), 223-227.

Näätänen, R., Kujala, T., & Winkler, I. (2011). Auditory processing that leads to conscious perception: a unique window to central auditory processing opened by the mismatch negativity and related responses. *Psychophysiology, 48*(1), 4-22.

NEVEN, A. (2015). *Explaining activity-related travel behaviour in persons with disabilities by means of health condition and contextual factors.* (PhD), (http://hdl.handle.net/1942/20558)

Ng, D. W. K., Soh, Y. W., & Goh, S. Y. (2014, 9-12 Dec. 2014). *Development of an Autonomous BCI Wheelchair.* Paper presented at the 2014 IEEE Symposium on Computational Intelligence in Brain Computer Interfaces (CIBCI).

Ngo, B.-V., & Nguyen, T.-H. (2022). A Semi-Automatic Wheelchair with Navigation Based on Virtual-Real 2D Grid Maps and EEG Signals. *Applied Sciences, 12*(17), 8880.

Omrani, M., Kaufman, M. T., Hatsopoulos, N. G., & Cheney, P. D. (2017). Perspectives on classical controversies about the motor cortex. *J Neurophysiol, 118*(3), 1828-1848.

Pakkenberg, B., Pelvig, D., Marner, L., Bundgaard, M. J., Gundersen, H. J. G., Nyengaard, J. R., & Regeur, L. (2003). Aging and the human neocortex. *Experimental gerontology, 38*(1-2), 95-99.

Pandarinath, C., Gilja, V., Blabe, C. H., Nuyujukian, P., Sarma, A. A., Sorice, B. L., Eskandar, E. N., Hochberg, L. R., Henderson, J. M., & Shenoy, K. V. (2015). Neural population dynamics in human motor cortex during movements in people with ALS. *eLife, 4*, e07436.

Plebe, A., & Grasso, G. (2019). The Unbearable Shallow Understanding of Deep Learning. *Minds and Machines, 29*(4), 515-553.

Polich, J. (2007). Updating P300: an integrative theory of P3a and P3b. *Clinical Neurophysiology, 118*(10), 2128-2148.

Pouget, A., Beck, J. M., Ma, W. J., & Latham, P. E. (2013). Probabilistic brains: knowns and unknowns. *Nature Neuroscience, 16*(9), 1170-1178.

Prathibanandhi, K., Selvapriya, V., Leela, M. K., Sivaprasad, R., Malini, V., & Kothai, R. (2022, 8-9 Dec. 2022). *Hand Gesture Controlled Wheelchair.* Paper presented at the 2022 International Conference on Power, Energy, Control and Transmission Systems (ICPECTS).

Proske, U., & Gandevia, S. C. (2012). The proprioceptive senses: their roles in signaling body shape, body position and movement, and muscle force. *Physiol Rev, 92*(4), 1651-1697.

Quax, S. C., Bosch, S. E., Peelen, M. V., & van Gerven, M. A. (2021). Population codes of prior knowledge learned through environmental regularities. *Scientific Reports, 11*(1), 640.

Remillard, E. T., Campbell, M. L., Koon, L. M., & Rogers, W. A. (2022). Transportation challenges for persons aging with mobility disability: Qualitative insights and policy implications. *Disability and Health Journal, 15*(1, Supplement), 101209.

Riesenhuber, M., & Poggio, T. (1999). Hierarchical models of object recognition in cortex. *Nature Neuroscience, 2*(11), 1019-1025.

Rohe, T., Ehlis, A.-C., & Noppeney, U. (2019). The neural dynamics of hierarchical Bayesian causal inference in multisensory perception. *Nature Communications, 10*(1), 1907.

Rolls, E. (2023). *Brain Computations and Connectivity*. Oxford: Oxford University Press.

Rumelhart, D. E., Hinton, G. E., & Williams, R. J. (1985). *Learning internal representations by error propagation*. Retrieved from

Saibene, A., Ghaemi, H., & Dagdevir, E. (2024). Deep learning in motor imagery EEG signal decoding: A Systematic Review. *Neurocomputing, 610*, 128577.

Schirrmeister, R., Springenberg, J., Fiederer, L., Glasstetter, M., Eggensperger, K., Tangermann, M., Hutter, F., Burgard, W., & Ball, T. (2017). Deep learning with convolutional neural networks for EEG decoding and visualization: Convolutional Neural Networks in EEG Analysis. *Human Brain Mapping, 38*.

Scott, S. H. (2008). Inconvenient truths about neural processing in primary motor cortex. *J Physiol, 586*(5), 1217-1224.

Serre, T., Wolf, L., Bileschi, S., Riesenhuber, M., & Poggio, T. (2007). Robust object recognition with cortex-like mechanisms. *Ieee Transactions on Pattern Analysis and Machine Intelligence, 29*(3), 411-426.

Shadmehr, R., Smith, M. A., & Krakauer, J. W. (2010). Error correction, sensory prediction, and adaptation in motor control. *Annual Review of Neuroscience, 33*(1), 89-108.

Shadmehr, R., & Wise, S. P. (2004). *The computational neurobiology of reaching and pointing: a foundation for motor learning*: MIT press.

Shapiro, L., & Spaulding, S. (2024). *Embodied Cognition*: Metaphysics Research Lab, Stanford University.

Shen, S., Cheng, J., Liu, Z., Tan, J., & Zhang, D. (2025). Bayesian inference-assisted reliability analysis framework for robotic motion systems in future factories. *Reliability Engineering & System Safety, 258*, 110894.

Silversmith, D. B., Abiri, R., Hardy, N. F., Natraj, N., Tu-Chan, A., Chang, E. F., & Ganguly, K. (2021). Plug-and-play control of a brain–computer interface through neural map stabilization. *Nature biotechnology, 39*(3), 326-335.

Sobierajewicz, J., Przekoracka-Krawczyk, A., Jaśkowski, W., Verwey, W. B., & van der Lubbe, R. (2017). The influence of motor imagery on the learning of a fine hand motor skill. *Exp Brain Res, 235*(1), 305-320.

Sobierajewicz, J., Szarkiewicz, S., Przekoracka-Krawczyk, A., Jaśkowski, W., & van der Lubbe, R. (2016). To What Extent Can Motor Imagery Replace Motor Execution While Learning a Fine Motor Skill? *Adv Cogn Psychol, 12*(4), 179-192.

Song, Y., Zheng, Q., Liu, B., & Gao, X. (2022). EEG conformer: Convolutional transformer for EEG decoding and visualization. *Ieee Transactions on Neural Systems and Rehabilitation Engineering, 31*, 710-719.

Tabar, Y. R., & Halici, U. (2017). A novel deep learning approach for classification of EEG motor imagery signals. *Journal of Neural Engineering, 14*(1).

Thomas Parr, G. P., Karl J. Friston. (2022). *Active Inference: The Free Energy Principle in Mind, Brain, and Behavior*.

Todorov, E. (2004). Optimality principles in sensorimotor control. *Nature Neuroscience, 7*(9), 907-915.

Turing, A. M. (2007). Computing machinery and intelligence. In *Parsing the Turing test: Philosophical and methodological issues in the quest for the thinking computer* (pp. 23-65): Springer.

Varela, F. J., Rosch, E., & Thompson, E. (1991). *The Embodied Mind: Cognitive Science and Human Experience*: The MIT Press.

Viviani, P., & Schneider, R. (1991). A developmental study of the relationship between geometry and kinematics in drawing movements. *Journal of experimental psychology: human perception and performance, 17*(1), 198.

Volkova, K., Lebedev, M. A., Kaplan, A., & Ossadtchi, A. (2019). Decoding Movement From Electrocorticographic Activity: A Review. *Front Neuroinform, 13*, 74.

Wan, Z., Li, M., Liu, S., Huang, J., Tan, H., & Duan, W. (2023). EEGformer: A transformer–based brain activity classification method using EEG signal. *Frontiers in Neuroscience, 17*, 1148855.

Wang, H., Li, Y., Long, J., Yu, T., & Gu, Z. (2014). An asynchronous wheelchair control by hybrid EEG-EOG brain-computer interface. *Cogn Neurodyn, 8*(5), 399-409.

Widyotriatmo, A., Suprijanto, & Andronicus, S. (2015, 31 May-3 June 2015). *A collaborative control of brain computer interface and robotic wheelchair.* Paper presented at the 2015 10th Asian Control Conference (ASCC).

Williams, T., & Scheutz, M. (2017). The state-of-the-art in autonomous wheelchairs controlled through natural language: A survey. *Robotics and Autonomous Systems, 96*, 171-183.

Willsey, M. S., Shah, N. P., Avansino, D. T., Hahn, N. V., Jamiolkowski, R. M., Kamdar, F. B., Hochberg, L. R., Willett, F. R., & Henderson, J. M. (2025). A high-performance brain–computer interface for finger decoding and quadcopter game control in an individual with paralysis. *Nature Medicine, 31*(1), 96-104.

Wolpert, D. M., Diedrichsen, J., & Flanagan, J. R. (2011). Principles of sensorimotor learning. *Nature Reviews Neuroscience, 12*(12), 739-751.

Wolpert, D. M., & Landy, M. S. (2012). Motor control is decision-making. *Current Opinion in Neurobiology, 22*(6), 996-1003.

Wu, W., Wu, C., Gao, S., Liu, B., Li, Y., & Gao, X. (2014). Bayesian estimation of ERP components from multicondition and multichannel EEG. *Neuroimage, 88*, 319-339.

Xie, Y., & Li, X. (2015). *A brain controlled wheelchair based on common spatial pattern.* Paper presented at the 2015 International symposium on bioelectronics and bioinformatics (ISBB).

Xiong, M., Hotter, R., Nadin, D., Patel, J., Tartakovsky, S., Wang, Y., Patel, H., Axon, C., Bosiljevac, H., Brandenberger, A., Bulger, M., Chien, W., Doyle, A., Hao, W., Jiang, J., Kim, K., Lahlou, S., Leung, C., Mamdani, M. H., Krzisnik, D. M., Marshall, K., Martin, T., Martinez, A., Mason, W., Odabassian, M., Riachi, R., Tan, J., Vosburg, S., Young, L., Yuan, J. M., & Zhen, A. (2019, 6-9 Oct. 2019). *A Low-Cost, Semi-Autonomous Wheelchair Controlled by Motor Imagery and Jaw Muscle Activation.* Paper presented at the 2019 IEEE International Conference on Systems, Man and Cybernetics (SMC).

Xu, J., Zheng, H., Wang, J., Li, D., & Fang, X. (2020). Recognition of EEG signal motor imagery intention based on deep multi-view feature learning. *Sensors, 20*(12), 3496.

Xu, Y., Wang, Q., Lillie, J., Kamat, V., & Menassa, C. (2025). CoNav Chair: Design of a ROS-based Smart Wheelchair for Shared Control Navigation in the Built Environment. *arXiv preprint arXiv:2501.09680*.

Yamins, D. L., & DiCarlo, J. J. (2016). Using goal-driven deep learning models to understand sensory cortex. *Nature Neuroscience, 19*(3), 356-365.

Yildiz, G., Duru, A. D., Ademoglu, A., Demiralp, T., & Ieee. (2007, Aug 22-26). *Bayesian EEG dipole source localization using SA-RJMCMC on realistic head model.* Paper presented at the 29th Annual International Conference of the IEEE-Engineering-in-Medicine-and-Biology-Society, Lyon, FRANCE.

Yu, Y., Zhou, Z., Liu, Y., Jiang, J., Yin, E., Zhang, N., Wang, Z., Liu, Y., Wu, X., & Hu, D. (2017). Self-Paced Operation of a Wheelchair Based on a Hybrid Brain-Computer Interface Combining Motor Imagery and P300 Potential. *Ieee Transactions on Neural Systems and Rehabilitation Engineering, 25*(12), 2516-2526.

Zhang, K., Xu, G., Zheng, X., Li, H., Zhang, S., Yu, Y., & Liang, R. (2020). Application of Transfer Learning in EEG Decoding Based on Brain-Computer Interfaces: A Review. *Sensors, 20*(21).

Zhao, W., Lu, H. D., Zhang, B. C., Zheng, X. W., Wang, W. F., & Zhou, H. F. (2025). TCANet: a temporal convolutional attention network for motor imagery EEG decoding. *Cognitive Neurodynamics, 19*(1).

Zhou, X., & Liao, P.-C. (2023). EEG-Based Performance-Driven Adaptive Automated Hazard Alerting System in Security Surveillance Support. *Sustainability, 15*(6), 4812.